\documentclass{ieeetj}
\usepackage{cite}
\usepackage{amsmath,amssymb,amsfonts}
\usepackage{algorithmic}
\usepackage{graphicx,color}
\usepackage{textcomp}
\usepackage{xcolor}
\usepackage{hyperref}
\hypersetup{hidelinks=true}

\usepackage{moreverb}
\usepackage{graphicx}
\usepackage{textcomp}
\usepackage{lipsum}
\usepackage{multirow}
\usepackage{algorithm}
\usepackage{amssymb}
\usepackage{threeparttable}
\usepackage{xcolor}
\usepackage{gensymb}
\usepackage{caption}
\usepackage{tabularx}
\let\labelindent\relax
\usepackage{enumitem}
\usepackage{adjustbox}

\usepackage{float}

\newcommand{\tb}[1]{\textbf{#1}}
\newcommand{\tu}[1]{\underline{#1}}

\def\BibTeX{{\rm B\kern-.05em{\sc i\kern-.025em b}\kern-.08em
    T\kern-.1667em\lower.7ex\hbox{E}\kern-.125emX}}
\AtBeginDocument{\definecolor{tmlcncolor}{cmyk}{0.93,0.59,0.15,0.02}\definecolor{NavyBlue}{RGB}{0,86,125}}

\def\OJlogo{\vspace{-4pt}$ $$ $}
\def\seclogo{\vspace{10pt}$ $$ $}

\def\authorrefmark#1{\ensuremath{^{\textbf{#1}}}}

\begin{document}
\receiveddate{XX Month, XXXX}
\reviseddate{XX Month, XXXX}
\accepteddate{XX Month, XXXX}
\publisheddate{XX Month, XXXX}
\currentdate{XX Month, XXXX}
\doiinfo{XXXX.XXXX.XXXXXXX}

\markboth{Drift-Corrected Monocular VIO and Perception-Aware Planning for Autonomous Drone Racing}{Azhari {et al.}}

\title{Drift-Corrected Monocular VIO and Perception-Aware Planning for Autonomous Drone Racing}
\author{Maulana Bisyir Azhari\authorrefmark{1}, Donghun Han\authorrefmark{1}, Je In You\authorrefmark{1}, Sungjun Park\authorrefmark{1}, and David Hyunchul Shim\authorrefmark{1}}
\affil{Department of Electrical Engineering, Korea Advanced Institute of Science and Technology (KAIST), Daejeon, 34141, South Korea}
\corresp{Corresponding author: David Hyunchul Shim (email: hcshim@kaist.ac.kr).}
\authornote{This work is supported by Institute of Civil Military Technology Cooperation funded by the Defense Acquisition Program Administration and Ministry of Trade, Industry and Energy of Korean government, Grant/Award Number: UM22206RD2}

\begin{abstract}
The Abu Dhabi Autonomous Racing League(A2RL) x Drone Champions League competition(DCL) requires teams to perform high-speed autonomous drone racing using only a single camera and a low-quality inertial measurement unit---a minimal sensor set that mirrors expert human drone racing pilots. 
This sensor limitation makes the system susceptible to drift from Visual-Inertial Odometry (VIO), particularly during long and fast flights with aggressive maneuvers. 
This paper presents the system developed for the championship, which achieved a competitive performance. 
Our approach corrected VIO drift by fusing its output with global position measurements derived from a YOLO-based gate detector using a Kalman filter. 
A perception-aware planner generated trajectories that balance speed with the need to keep gates visible for the perception system. 
The system demonstrated high performance, securing podium finishes across multiple categories: third place in the AI Grand Challenge with top speed of 43.2 km/h, second place in the AI Drag Race with over 59 km/h, and second place in the AI Multi-Drone Race. 
We detail the complete architecture and present a performance analysis based on experimental data from the competition, contributing our insights on building a successful system for monocular vision-based autonomous drone flight.
\end{abstract}


\maketitle

\section{INTRODUCTION}
\label{sec:introduction}
\IEEEPARstart{A}{utonomous} drone racing (ADR) competitions\cite{moon2017iros_competitions, jung2018direct_visual_servoing, moon2019challenges_adr, alphapilot, foehn2022alphapilot_uzh, de2022mavlab_airr, a2rl_x_dcl} have rapidly emerged as a premier testbed for advancing the fields of aerial robotics and physical artificial intelligence, challenging researchers to push the known boundaries of perception, planning, and control within highly dynamic and unforgiving environments. 
In this setting, autonomous drones must navigate complex, three-dimensional racecourses at extreme speeds, executing a sequence of precise and aggressive maneuvers while passing through gates, all under severe time and computational constraints\cite{hanover2024adr_survey}.

\begin{figure*}[t!]
    \centering
    \includegraphics[width=1.0\textwidth]{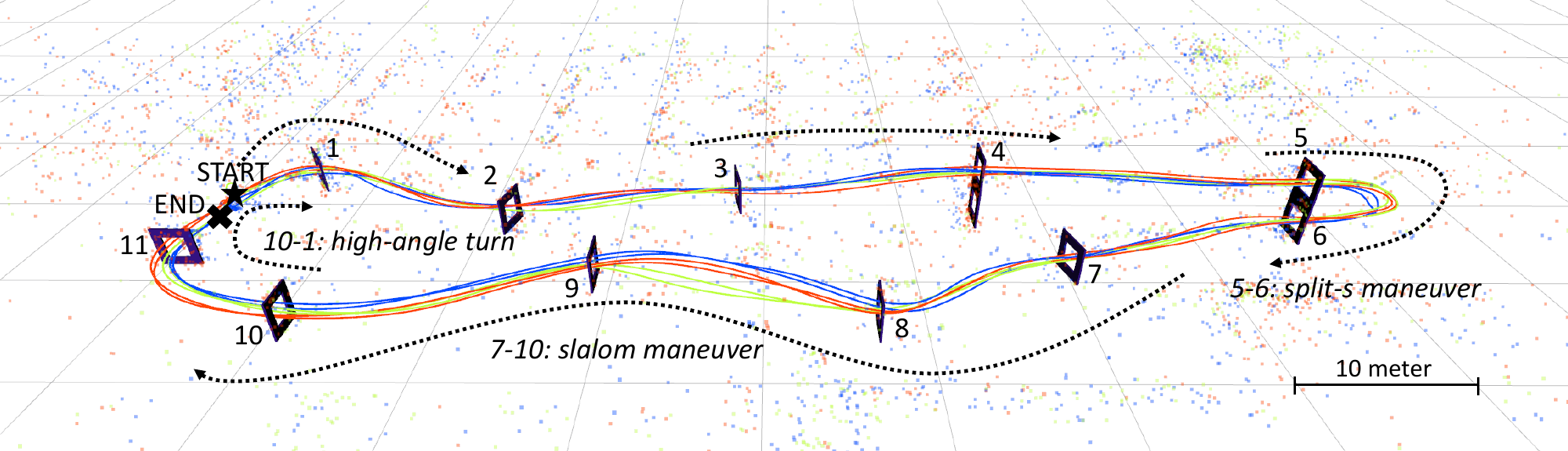}
    \caption{Visualization of the drone's trajectories across multiple high-speed runs on the final A2RL$\times$DCL championship track. The colored lines represent distinct race trajectories, showing tight clustering even through aggressive maneuvers. The numbers represent the gate ordering, while the annotations highlight the track's complexity, including a split-s maneuver (gates 5-6), a slalom (gates 7-10), and a high-angle turn (gates 10-1). These trajectories are reconstructed post-flight using MAPLAB~\cite{schneider2018maplab, cramariuc2022maplab}, which serve as the ground truth for our system's evaluation. The colored points are the visual landmarks used in MAPLAB framework.}
    \label{fig:team_kaist_final_runs}
    \vspace{-8pt}
\end{figure*}

The Abu Dhabi Autonomous Racing League x Drone Champions League (A2RL$\times$DCL)\cite{a2rl_x_dcl} represents a significant milestone in this domain.
The championship's primary difficulty stems from its strict sensory limitations.
Competing teams are restricted to using only a single forward-facing RGB camera and optionally a low-cost inertial measurement unit (IMU) built in the flight controller for all navigation and control tasks.
This minimal sensor set is intentionally designed to mirror the configuration used by expert human first-person view (FPV) pilots, who use a single camera feed and rely on their visual-motor coordination skills for flight stabilization and control\cite{pfeiffer2021human}.

This constraint is a substantial departure from previous events that permitted more extensive sensor arrays, which may include stereo cameras, RGB-D camera, or even LiDAR, to perceive the environment.
By limiting the perception system to a single monocular camera, the championship presents a more profound scientific challenge. 
The core of this challenge lies in overcoming the inherent limitations of Visual-Inertial Odometry (VIO), the standard technique for estimating a drone's pose and velocity from a camera and IMU.
While effective in less demanding scenarios, VIO is notoriously susceptible to accumulating significant drift during the long, aggressive, high-speed flights characteristic of drone racing, where perceptual challenges such as motion blur and sparse features are common.

This paper presents the developed system, which secured third place by integrating robust gate-corrected state estimation and perception-aware trajectory planning. 
A visualization of multiple high-speed race runs flown by our drone on the championship track is shown in Figure \ref{fig:team_kaist_final_runs}.

\subsection{Competition Overview}
The Abu Dhabi Autonomous Racing League in collaboration with the Drone Champions League has established a new benchmark for autonomous flight. 
The competition challenges teams to race high-speed drones through complex aerial courses using a minimal sensor suite: a single forward-facing RGB camera and an IMU. 
This intentional limitation poses a significant test of perception-based autonomy, particularly for high-speed drone flight in potentially visually sparse environments. 
The drones, capable of speeds exceeding 100 km/h, must navigate a series of identical 1.5$\times$1.5 m$^2$ gates in a 95$\times$28 m$^2$ area.

\begin{figure*}[t!]
    \centering
    \includegraphics[width=1.0\textwidth]{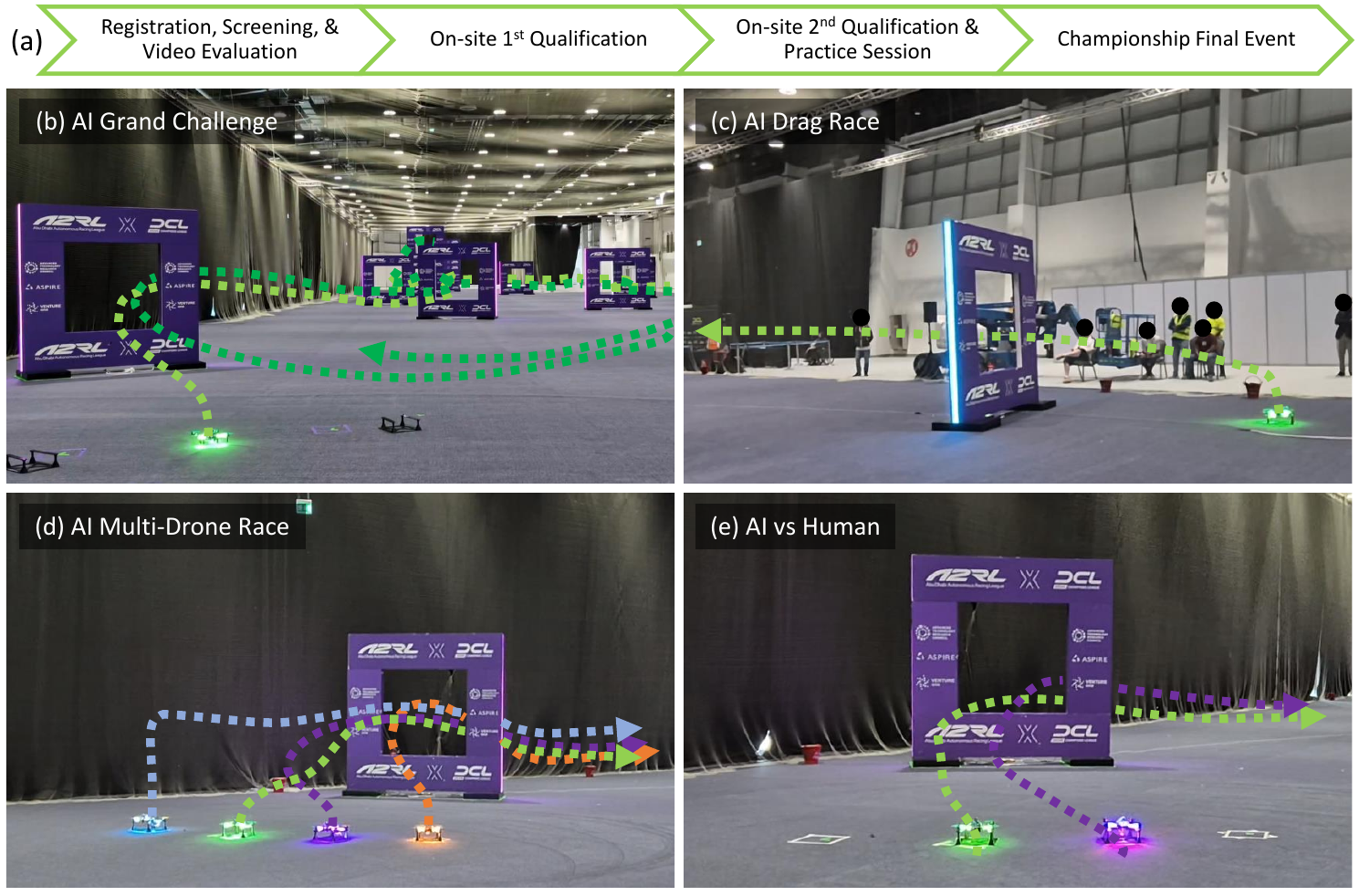}
    \caption{(a) Championship stages. (b)-(e) The four distinct race categories featured in the A2RL$\times$DCL Autonomous Drone Championship.}
    \label{fig:a2rl_x_dcl_championship}
    \vspace{-8pt}
\end{figure*}

As illustrated in Figure \ref{fig:a2rl_x_dcl_championship}, the championship is a multi-stage event that progresses from initial qualification rounds to the final race day (Figure \ref{fig:a2rl_x_dcl_championship}a). 
It features four distinct race categories designed to test different facets of autonomous flight:
(a) \textbf{AI Grand Challenge}---A two-lap, single-drone time trial on a 170-meter track, serving as the primary test of overall system performance (Figure \ref{fig:a2rl_x_dcl_championship}b); (b) \textbf{AI Drag Race}---A straight-line challenge focused on maximizing raw speed and maintaining high-speed stability (Figure \ref{fig:a2rl_x_dcl_championship}c); (c) \textbf{Multi-Drone Race}---A four-drone simultaneous race designed to test collision avoidance and strategic racing line optimization (Figure \ref{fig:a2rl_x_dcl_championship}d); and (d) \textbf{AI vs Human}---A head-to-head duel pitting the top AI-powered drones against elite human FPV pilots (Figure \ref{fig:a2rl_x_dcl_championship}e).

\subsection{Contributions}
This paper presents the design, implementation, and performance analysis of the system developed for the A2RL$\times$DCL Championship. Our primary contributions are:
\begin{itemize}
    \item \textbf{System Architecture}: A holistic and competition-validated architecture for monocular drone racing, presenting key design principles for successfully integrating state-of-the-art perception, estimation, planning, and control modules. 
    \item \textbf{Competition Results}: A detailed analysis of the system's performance, which achieved a top speed of 43.2 km/h in the AI Grand Challenge and over 59 km/h in the AI Drag Race, securing podium finishes in multiple categories.
    \item \textbf{Discussion and Lessons Learned}: A discussion of the system's limitations, key insights gained from the competition, and potential directions for future research in high-speed autonomous flight.
\end{itemize}

\subsection{Paper Structure}
The remainder of this paper is organized as follows. Section \ref{sec:related_work} reviews related work in autonomous drone racing. Section \ref{sec:dev_principle_strategy} outlines our core development principles and strategy. 
Section \ref{sec:impementation} details the hardware and software implementation of each system module. 
Section \ref{sec:results} presents a thorough analysis of our experimental and competition results. 
Section \ref{sec:discussion} discusses the limitations, lessons learned, and future challenges.
Finally, Section \ref{sec:Conclusion} concludes the paper.

\section{RELATED WORK}
\label{sec:related_work}
This section reviews the strategies and state-of-the-art approaches developed within each of these core modules in the context of ADR~\cite{hanover2024adr_survey}.

\subsection{Perception and State Estimation}
Early attempts in ADR~\cite{moon2019challenges_adr}, which used established methods like Visual Odometry (VO) or feature-based SLAM (e.g., ORB-SLAM \cite{mur2015orb1, campos2021orb3}), proved to be insufficient. 
While extending these to Visual-Inertial Odometry (VIO) by fusing camera data with high-frequency IMU measurements helps resolve scale ambiguity and provides robustness against transient visual failures\cite{bloesch2017rovio, qin2018vinsmono, xiaochen2020larvio, geneva2020openvins}, the visual front-end still breaks down under the sustained perceptual degradation of a real race\cite{foehn2022alphapilot_uzh}. 

This led to a paradigm shift towards hybrid, ADR-specific systems that leverage deep learning for perception and classical filters for estimation\cite{li2020adr_vml, kazim2022perception_aware_adr}. 
This architecture, which dominated competitions such as the 2019 AlphaPilot Challenge, uses Convolutional Neural Networks (CNNs) to robustly detect race gates—-or their corners—-despite visual corruption\cite{de2022mavlab_airr, foehn2022alphapilot_uzh}. 
The precise location of these detected gates then serves as a corrective measurement within an Extended Kalman Filter (EKF) to eliminate the accumulated drift of the VIO subsystem, effectively anchoring the drone's state estimate to a known track map. 
Winning strategy optimized this pipeline by focusing on computational efficiency and robustness to sensor outliers, employing Moving Horizon Estimation (MHE)\cite{de2022mavlab_airr}.

Gate detection has been an integral part since the beginning of ADR competitions and has become more important.
Early works employed color-based\cite{jung2018direct_visual_servoing}, snake-gate\cite{li2020adr_mavlab}, and CNN-based gate detections\cite{jung2018perception_guidance_adr}.
Recent advances employ more sophisticated algorithms, such as segmentation\cite{de2022mavlab_airr, kazim2022perception_aware_adr, kaufmann2023champion}, heatmap\cite{foehn2022alphapilot_uzh, cao2017realtime_paf_detection, cao2019openpose_paf}, and direct gate pose detection\cite{Pham2021gatenet, qiao2024continual_gate_detection, pham2022pencilnet}.
There are several strategies to detect gates and estimate poses from it: (a) segmentation$\rightarrow$corner detection$\rightarrow$pose estimation\cite{de2022mavlab_airr, kazim2022perception_aware_adr}; (b) part segmentation$\rightarrow$corner association$\rightarrow$pose estimation\cite{foehn2022alphapilot_uzh, kaufmann2023champion}; (c) direct pose regression\cite{Pham2021gatenet, qiao2024continual_gate_detection, pham2022pencilnet}.
However, these methods are time-consuming to label\cite{de2022mavlab_airr, kazim2022perception_aware_adr}, complex, multi-stage corners detection\cite{foehn2022alphapilot_uzh, kaufmann2023champion}, and require exact 3D labels \cite{Pham2021gatenet, qiao2024continual_gate_detection, pham2022pencilnet}.

Another area of advancement is the direct estimation of the drone's state via deep learning from IMU data\cite{cioffi2023learned_inertial_odometry, qiu2025airio}, which integrate a model-based filter with a learning-based inertial odometry, showing superior performance in perceptually challenging racing scenarios compared to VIO baselines.
While promising, these methods require significant amount of data and present generalization and computation challenges, making them hard to deploy in a time-constrained competitive event.

\subsection{Planning and Control}
Early autonomous drone racing competitions saw a dichotomy in navigation strategies between visual servoing and path-following. 
Visual servoing approaches\cite{jung2018direct_visual_servoing, jung2018perception_guidance_adr, qin2023perception_aware_ibvs}, successful in early events like the IROS ADR competitions, directly use the detected gate's location in the image plane to generate control commands.
Over time, the dominant strategy has shifted to path or trajectory following \cite{moon2019challenges_adr, li2020adr_mavlab, foehn2022alphapilot_uzh, de2022mavlab_airr}.
This approach requires more accurate state estimation but allows for more complex, longer-horizon, and aggressive trajectory planning.

Early work in trajectory planning by \cite{Mellinger2011minsnap} exploited the differential flatness of quadrotor dynamics, as polynomials, enabling the generation of smooth, dynamically feasible maneuvers.
\cite{Mueller2015} generated computationally efficient motion primitives for rapid trajectory synthesis.
In ADR context, the AlphaPilot challenge saw the successful use of a sampling-based planner that randomly samples velocity states at upcoming gates and connects them with closed-form primitives~\cite{foehn2022alphapilot_uzh}.
The pursuit of maximum speed pushed the field from feasible to truly time-optimal planners. 
CPC~\cite{foehn2021CPC} proposed a formulation of progress allowing to simultaneously solve for the optimal trajectory and the time allocation. 
TOGT planner~\cite{qin2024togt} further improved performance by modeling gates as spatial volumes rather than single waypoints, allowing the planner to exploit the full flyable area to find faster trajectory.
Furthermore, TOGT requires significantly less optimization time compared to CPC.

Once a trajectory is generated, a high-performance controller is required for accurate tracking. 
While PID controllers are common, their performance degrades in aggressive flight.
Geometric controllers\cite{taeyoung2010geometric, taeyoung2013nonlinearrobustgeometric, Mellinger2011minsnap} can be a simple yet significant improvement compared to PID.
% Advanced model-based methods are therefore preferred. 
MPC is highly effective due to its predictive nature and ability to handle actuator constraints~\cite{sun2022comparative_nmpc_dfbc}. 
A specialized variant, MPCC\cite{romero2022mpcc, krinner2024mpcc++}, is particularly well-suited for racing as it optimizes progress along a geometric path rather than tracking a fixed time-parameterized trajectory. 

Learning-based methods have also gained significant traction. 
A notable achievement in this area is the use of deep reinforcement learning to train policies capable of champion-level performance\cite{kaufmann2023champion}, demonstrating that an end-to-end control approach can achieve highly competitive results.
\cite{song2023oc_vs_rl} showed that reinforcement learning can outperform optimal control.
Recent works\cite{geles2024demonstrating,xing2025bootstrapping} have shown the feasibility of end-to-end vision-based flights in ADR.
While promising, deploying learning-based method can be very challenging especially in time-limited events.

\begin{figure*}[h!]
\begin{center}
\includegraphics[width=0.9\textwidth]{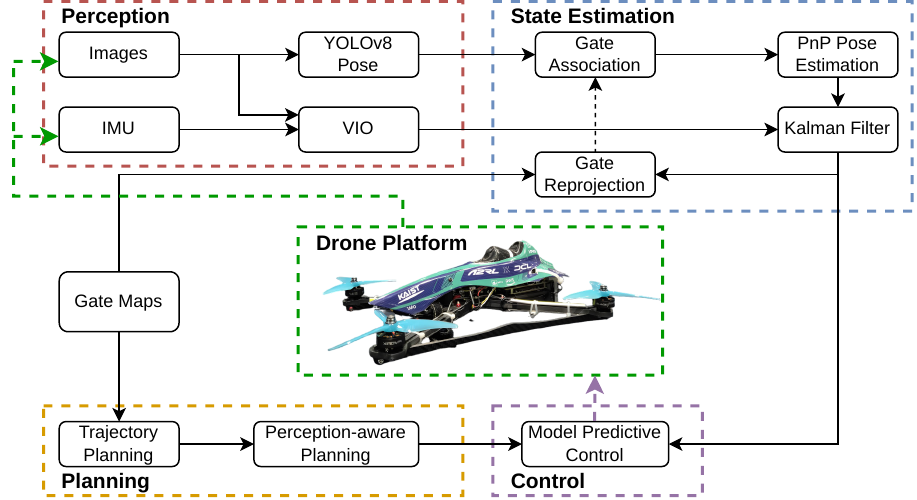}
\end{center}
\caption{Overview of the implemented system architecture for the A2RL$\times$DCL Championship. The system is composed of four main modules: Perception, State Estimation, Planning, and Control. The Perception module utilizes a YOLO-based gate keypoints detector and a VIO system to process incoming image and IMU data. The State Estimation module then fuses the VIO output with gate detection measurements via a Kalman Filter to produce a drift-corrected state estimate. Prior to the race, the Planning module generates a time-optimal trajectory to traverse the gates with perception-awareness in mind. Then the MPC module computes the commands for the drone platform in real-time.}
\label{fig/system_overview}
\end{figure*}

\section{DEVELOPMENT PRINCIPLE AND STRATEGY}
\label{sec:dev_principle_strategy}
Our development was founded on three core principles: \textbf{rapid development}, \textbf{ease of deployment}, and \textbf{iterative improvement},
which prioritized robustness and consistent performance. 
The goal of this approach was to produce a reliable system that could be deployed quickly during the limited preparation timeline and then refined with empirical data gathered during experiments. 
This refinement process involved either improving the already deployed/developed algorithms or upgrading to more advanced and higher-performing ones.
These principles were essential because the A2RL$\times$DCL championship was the first major autonomous drone racing event in five years since AlphaPilot\cite{alphapilot}. 
This meant that recent advancements in the field were largely lab-tested, making the rapid selection, validation, and adjustment of the right algorithms a critical factor for success.

The first focus was to select a navigation strategy appropriate for the competition. 
% While visual servoing was successful in earlier competitions\cite{jung2018direct_visual_servoing}, 
We chose the path-following strategy, which has been proven effective in more recent competitions\cite{de2022mavlab_airr, foehn2022alphapilot_uzh}. 
This approach, however, requires accurate state estimation to navigate gates successfully.
Recognizing that long, multi-lap races would inevitably cause significant drift in any standalone VIO system, we established drift-corrected state estimation as the foundational pillar of our architecture.
We further discuss the strategy for each module of the system.

Second, to achieve an accurate drift-corrected state estimation, we selected the most reliable monocular VIO algorithm to serve as the basis. 
Lacking an on-premise, large-scale motion-capture system, we benchmarked various open-source VIO algorithms on a public dataset that is similar to the drone's specifications and the racing context. 
We chose the subset of TII-RATM dataset\cite{bosello2024tii_ratm} for this evaluation.

Third, to correct the unavoidable drift in the VIO, we needed to provide a corrective measure based on globally known landmarks, which are the gates.
While prior methods have been proven successful, they often require a significant amount of time for labeling\cite{de2022mavlab_airr, kaufmann2023champion, kazim2022perception_aware_adr}.
Therefore, we opted for a more straightforward approach: corner keypoints detection based on YOLOv8 architecture\cite{maji_2022_yolo_pose, yolov8_ultralytics}.
Additionally, keypoints are also less time-consuming to label compared to segmentation.

Fourth, for the trajectory planning, we experimented with simpler methods, such as simple mathematical hard-coded path and polynomial-based\cite{Mellinger2011minsnap}. 
We then employed more advanced and high-performing algorithms such as PMM\cite{teissing2024realtime_planning_pmm} and TOGT\cite{qin2024togt}. 
The trajectory was then tracked with geometric control\cite{taeyoung2010geometric, sun2022comparative_nmpc_dfbc} or optimal-control MPC\cite{foehn2022agilicious}.
In the end, we employed the modified TOGT, introducing perception-awareness to enhance state-estimation reliability, which was then track by the MPC and low-level PID controller.
% \pagebreak

Ultimately, these strategic choices enabled a rapid and iterative development cycle. 
During the limited preparation timeframe and on-site practice sessions, each flight served as an opportunity for data collection, allowing us to incrementally augment our gate detection dataset, refine the state-estimation pipeline, adjust the planner, and improve controller performance. 
The overview of the final implemented system architecture for the competition can be seen in Figure \ref{fig/system_overview}.

\begin{figure}[t!]
    \centering
    \includegraphics[width=0.99\columnwidth]{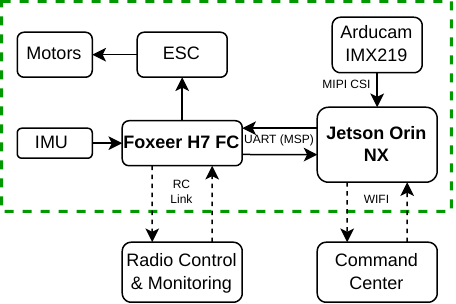}
    \caption{Hardware and Sensor Interfaces.}
    \label{fig:hardware_and_interfaces}
    \vspace{-12pt}
\end{figure}

\begin{figure*}[t!] % 'h'ere, 't'op, 'b'ottom, 'p'age of floats
    \centering % Center the figures
     \includegraphics[width=1.0\textwidth]{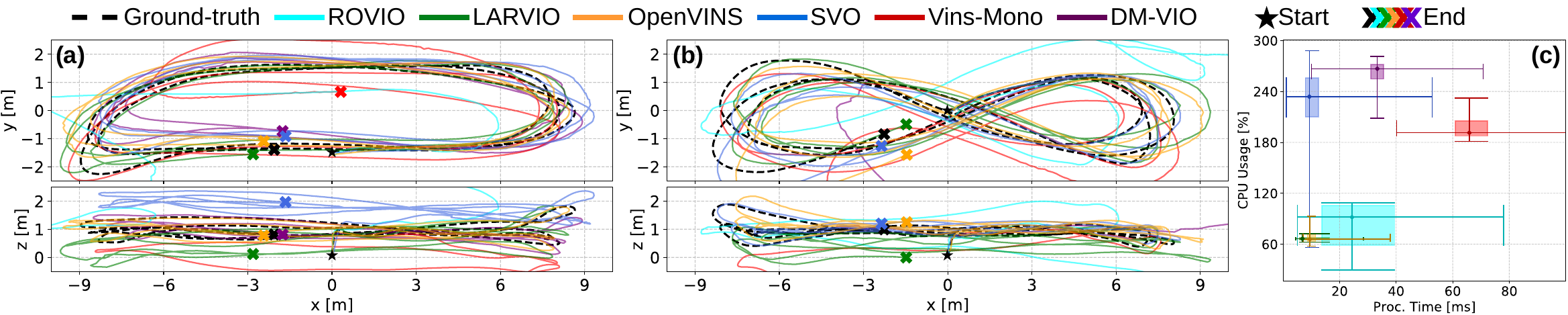}
    \caption{Visualization of preliminary comparison of VIO algorithms on TII-RATM piloted sequences. VIO trajectory visualization on TII-RATM (a) 01p and (b) 07p sequences, respectively. LARVIO, OpenVINS, and SVO perform well on both sequences; DM-VIO fails on sequence 07p; ROVIO and VINS-Mono fail on both sequences. (c) Statistics of processing time and CPU usage of the VIO algorithms on Jetson Orin NX.}
    \label{fig:tii_ratm_vio_comparison}
    \vspace{-16pt}
\end{figure*}

\begin{table*}[t!] % Placement specifiers: h-here, t-top, b-bottom, p-page of floats
    \centering % Center the table on the page
    \caption{Preliminary VIO comparisons on TII-RATM piloted sequences.}
    \label{tab:vio_tii_ratm} % Label for referencing the table
    \setlength{\tabcolsep}{8pt}
    \begin{tabular}{l|cccc|cccc|ccc} % Column definitions: l-left, c-center
        \hline\hline % Double horizontal line at the top
        % Header - Row 1
        \multirow{2}{*}{\textbf{Method}} &\multicolumn{4}{c}{\textbf{01p-ellipse}} &\multicolumn{4}{c}{\textbf{07p-lemniscate}} &\multicolumn{3}{c}{\textbf{Computation}}\\
        &ATE$_t$ &ATE$_r$ &t$_{rel}$ &r$_{rel}$ &ATE$_t$ &ATE$_r$ &t$_{rel}$ &r$_{rel}$ &time [ms] &CPU [\%] &Mem [MB] \\
        \hline % Single horizontal line separating header from data
        ROVIO &21.8 &15.0 &76.1 &2.08 &21.3 &118.6 &80.24 &2.38     
                &26.7 $\pm$ 15.4    &82.2 $\pm$ 25.7 &270.7 $\pm$ 0.7\\
        LARVIO &0.50 &\tb{1.80} &\tu{3.10} &\tb{0.98} &0.48 &4.42 &3.24 &\tu{1.12}      
                &\tb{10.3 $\pm$ 5.6}     &\tb{66.5 $\pm$ 1.9} &\tb{77.8 $\pm$ 3.5}   \\
        OpenVINS &\tb{0.29} &\tu{2.17} &\tb{2.23} &\tu{1.18} &\tu{0.38} &\tu{1.99} &\tu{2.49} &1.21    
                &\tu{11.4 $\pm$ 5.2}     &\tu{68.2 $\pm$ 7.2} &112.6 $\pm$ 1.4\\
        SVO &\tu{0.38} &3.17 &3.17 &1.24 &\tb{0.18} &\tb{1.56} &\tb{1.63} &\tb{1.10}         
                &11.5 $\pm$ 6.4     &230.5 $\pm$ 38.8 &181.6 $\pm$ 15.8\\
        VINS-Mono &5.23 &17.5 &22.7 &1.58 &4.98 &46.4 &18.8 &1.41   
                &67.7 $\pm$ 9.4     &197.5 $\pm$ 16.6 &\tu{89.3 $\pm$ 5.2}\\
        DM-VIO &0.34 &2.11 &2.79 &0.99 &52.4 &166.2 &236.6 &4.96    
                &33.3 $\pm$ 3.9     &261.7 $\pm$ 14.9 &290.7 $\pm$ 6.5\\
        \hline
        \hline % Double horizontal line at the bottom
        \multicolumn{12}{l}{\parbox{0.9\textwidth}{
        \begin{itemize}[leftmargin=*]
            \item[-] The CPU usage is the percentage of 1-cpu core usage, with the total of 800$\%$ for the total 8-core CPU.
            \item[-] The best and second-best values are \tb{bold} and \tu{underlined}, respectively.
        \end{itemize}
        }}
    \end{tabular}
    \vspace{-8pt}
\end{table*}

\section{IMPLEMENTATION}
\label{sec:impementation}

\subsection{Drone Hardware and Software}
\textbf{Physical Specifications.} Teams participating in the competition were given pre-made drones with a fixed hardware configuration, with the exception of the camera mounting direction.
The drone physical hardware consisted of 30$\times$28$\times$10 cm frame, 5 inch propellers, XNOVA Black Thunder 2207 motors with 2100KV, and a LiPo 6S 1400mAh battery, with a total weight approximately 960 grams.

\textbf{Computes.} The main compute was NVIDIA Jetson Orin NX on a A603 carrier board. 
It had 8-core ARM cortex A78AE, 16GB RAM, 1024-core NVIDIA Ampere GPU, and 1TB SSD storage, which was used for high-level computation and image processing obtained from Arducam 8MP IMX219 wide angle camera.
The Jetson computer was equipped with Ubuntu 20.04 based operating system, jetpack 5.1.2, cuda 11.4, and ROS Noetic.
A Foxeer H7 was used for flight controller (FC) computer, it was equipped with STM32H743 ARM 480MHZ CPU, MPU-6000 6-axis IMU, and interfaces to control the motors, WIFI, radio-control (RC) receiver as well as video streaming.
The FC was flashed with Betaflight firmware version 4.4.0.
The Jetson and FC communicate via a UART interface with MultiWii Serial Protocol (MSP).
The main data being transferred are the IMU (250Hz) and control command (200Hz), while other data are optional.

% 3280x2464@21fps, 3280x1848@28fps, 1920x1080@30fps, 1640x1232@30fps, 1280x720@60fps
\textbf{Camera.} The only exteroceptive sensor used is an Arducam RGB rolling shutter camera with Sony IMX219 sensor and wide-angle lens. 
It had 155$^\circ$(H)$\times$115$^\circ$(V)$\times$175$^\circ$(D) field-of-view (FOV) and several video streaming modes: (i) 3280$\times$2464 @ 21FPS; (ii) 3280$\times$1848 @ 28FPS; (iii) 1920$\times$1080 @ 30FPS; (iv) 1640$\times$1232 @ 30FPS; and (v) 1280$\times$720 @ 30FPS.
We chose to use the mode (iv) as the mode (v) will result in a reduced vertical FOV.
The image was then downscaled by a factor of 2 to reduce the computation, resulting in an image stream of 820$\times$626 @ 30FPS.
It was connected to an Orin NX computer through MIPI-CSI-2 interface.
The camera can be mounted from 15$^{\circ}$-55$^{\circ}$ pitch angle, we mounted the camera at 15$^\circ$ for the qualification and increased it to around 45$^{\circ}$ as we increased the speed for the final event.
We set a fixed exposure time and gain to the camera to enforce lighting consistency and reduce unwanted artifacts due to automatic camera control.
The camera intrinsics and extrinsics were estimated using Kalibr\cite{furgale2013kalibr}.

\textbf{Sensor Timings.} Accurate sensor timestamps are critical for the performance of VIO, yet the drone platform lacks hardware-level synchronization between the camera and IMU. 
To overcome this, we implemented specific software-based timing measures. 
For the camera, we avoided standard APIs that timestamp frames upon their arrival in the processing queue, as this can introduce non-deterministic latency. 
Instead, we developed a custom GStreamer pipeline within a ROS node to capture not only the image but also its precise hardware acquisition timestamp. 
% This approach provides a much more accurate and stable time reference for when the image was actually exposed. 
For the IMU, we observed a significant time-drift between the FC and the Jetson clocks. 
Therefore, we opted to use the IMU packet's arrival time on the Jetson. 
While this method introduces some minor, high-frequency jitter, it is more reliable than a drifting timestamp from the FC.
The final, constant time offset between these two was estimated using the Kalibr toolbox\cite{furgale2013kalibr}.

\subsection{Perception}
\label{sec:perception}
The perception module processed raw image input and IMU data into positional information which can be utilized for the state-estimation module. 
It consisted of visual-inertial odometry (VIO) to estimate the drone's pose given the surrounding features.
Then, a gate detection module estimated the gate's location relative to the drone.

\subsubsection{Visual-Inertial Odometry Algorithms}
\label{sec:perception_vio}
Visual-Inertial Odometry (VIO) fuses data from camera and Inertial Measurement Unit (IMU), offering a compelling solution with minimal sensors requirement, making it suitable for agile racing platforms. 
However, the extreme dynamics inherent in drone racing—characterized by high speeds, aggressive accelerations, and rapid rotations—pose significant challenges. 
These include severe motion blur, potential texture scarcity on tracks, and stringent demands on computational efficiency and low-latency processing for real-time control. 
Furthermore, practical deployment necessitates robust initialization procedures and resilience to sensor calibration inaccuracies. 
Therefore, selecting an appropriate VIO algorithm requires careful consideration of these factors.
Based on the available literature, we selected several VIO algorithms as candidates that will be used in our system: ROVIO\cite{bloesch2017rovio}, LARVIO\cite{xiaochen2020larvio}, OpenVINS\cite{geneva2020openvins}, SVO\cite{forster2016svo}, Vins-Mono\cite{qin2018vinsmono}, and DM-VIO\cite{von2022dmvio}.
To benchmark these VIOs, we used the RMS of Absolute Trajectory Error for both the translation (ATE$_t$, m) and rotation (ATE$_r$, deg)\cite{zhang2018tutorial_vio_eval, grupp2017evo} parts, as well as the RMS of Relative Pose Error (RPE) for the both the translation ($t_{rel}$, \%) and rotation ($r_{rel}$, deg/10m) parts along multiple segments $\Delta d \in \{5,10,...,50\}$ meters\cite{geiger2012kitti_dataset, grupp2017evo}.

In order to choose the VIO algorithm for the competition, we conducted preliminary experiments on publicly available dataset, specifically TII-RATM dataset\cite{bosello2024tii_ratm}. 
We chose TII-RATM dataset due to its closest similarities in terms of drone's specification, such as the camera and IMU sensors, compared to UZH-FPV\cite{delmerico2019uzh_fpv_dataset} and BlackBird\cite{amado2020blackbirddataset} Datasets. 
However, TII-RATM does not provide the exact camera-imu extrinsics calibration, which is very important for VIO algorithms. 
To resolve this issue, we ran several VIO algorithms that are able to do online extrinsics calibration and try to use the calibrated extrinsics to other VIO methods. 
We found that the result from OpenVINS\cite{geneva2020openvins} was the most accurate and reliable, which can be used by other VIO algorithms. 
For evaluation, we dropped 3 out of 4 camera frames, such that the resulting camera frequency used is 30hz. 
The IMU frequency is 500hz. 
To choose the most suitable VIO method, we used the following criteria: accuracy, robustness, processing time, cpu usage, and memory usage, in order of the most to less important. 
For hardware and time related criteria, we set maximum thresholds which should be met: 33 ms for processing time, 200\% CPU usage, and 500 MB memory usage.

The preliminary VIO comparisons can be seen in Figure \ref{fig:tii_ratm_vio_comparison} and Table \ref{tab:vio_tii_ratm}. 
For the sake of conciseness, we only evaluated the VIO algorithms on the first 3 laps of sequence 01p-ellipse and 07p-lemniscate. 
Note that we were unable to perform evaluation on the autonomous sequences due to insufficient IMU excitation needed for VIO initialization at the start of the sequence. 
In terms of accuracy and robustness, LARVIO, OpenVINS, and SVO were superior compared to ROVIO, VINS-Mono, and DM-VIO. 
One can argue that the inferior accuracies of ROVIO, VINS-Mono, and DM-VIO are due to inaccuracies of camera-imu extrinsics values, however, the other methods have shown to be robust to these inaccuracies.
Out of the top 3 methods, SVO required significantly higher CPU usage compared to LARVIO and OpenVINS.
Between LARVIO and OpenVINS, we found that OpenVINS was more robust during initialization, especially due to its persistent "SLAM features" which prevented it from drifting-away when there was no motion. 
While LARVIO has similar functionality leveraging its zero-velocity update (ZUPT) in case no motion has appeared, we observed that it was less robust compared to OpenVINS's implementation.
Based on these results, we selected OpenVINS as the VIO backbone for the competition.

\begin{figure}[t!]
    \centering
    \includegraphics[width=\columnwidth]{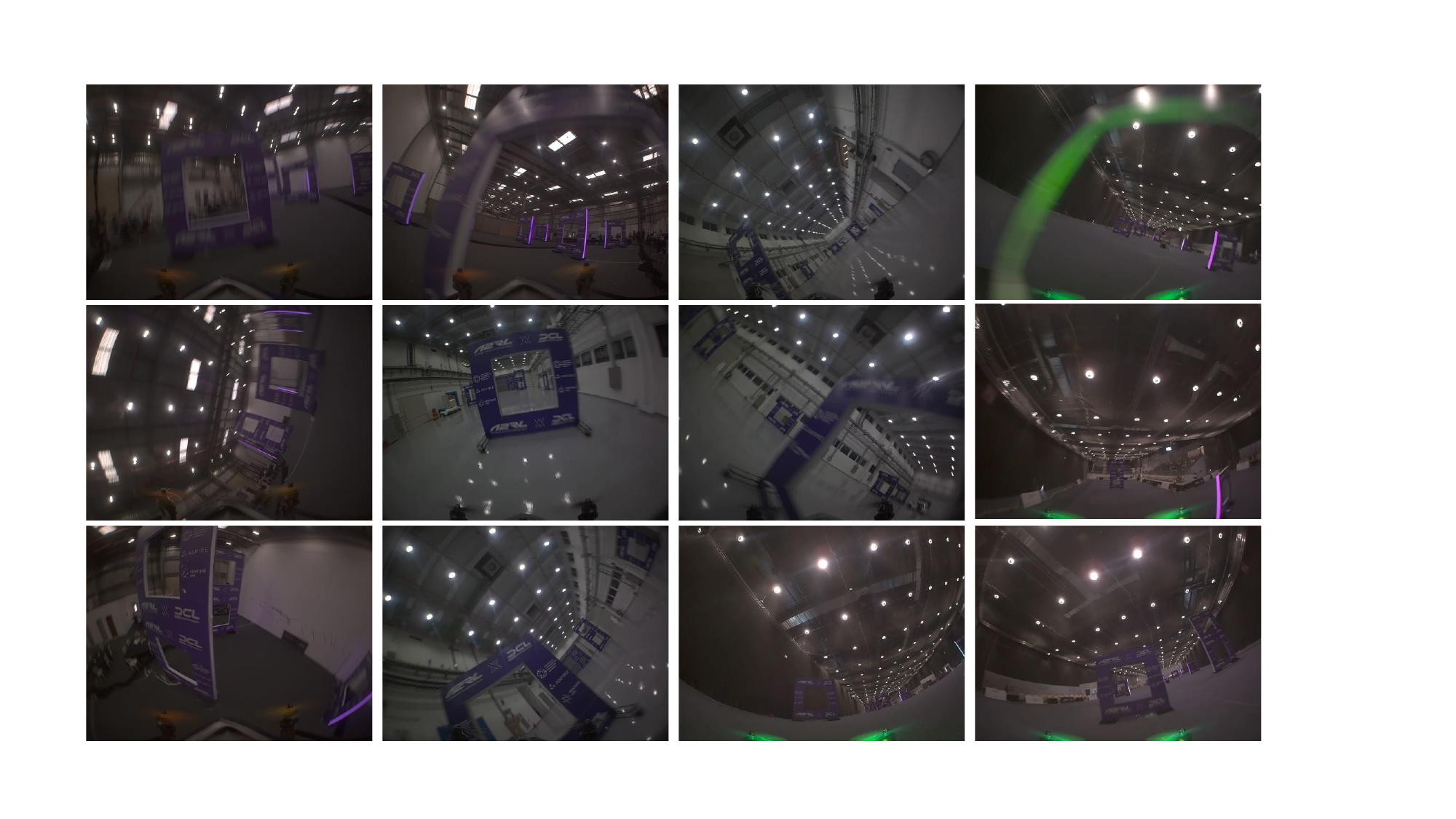}
    \caption{Examples of gate detection dataset collected incrementally prior and during the competition. The images were then resized into 640$\times$640 pixels for inference.}
    \label{fig:gate_dataset}
\end{figure}

\subsubsection{Gate Corner Detection with YOLOv8-Pose}
Unlike prior two-stage methods~\cite{de2022mavlab_airr,foehn2022alphapilot_uzh}, our perception module utilized YOLO-Pose\cite{maji_2022_yolo_pose} for simultaneous gate detection and corner keypoint localization. 
This single-pass approach enhanced speed, making it suitable for real-time application. 
Additionally, the annotation process of bounding boxes and keypoints of each gate was less labor-intensive than segmentation mask labeling required by alternative methods, a significant advantage under competition time constraints.

Our detection module utilizes the YOLOv8-Pose~\cite{yolov8_ultralytics} object detection model, predicting bounding boxes and inner gate corner keypoints concurrently. Following ~\cite{bosello2024tii_ratm}, the network outputted labels comprising:
\begin{align}
% [O, c_x,c_y,w,h,tl_x,tl_y,tl_v,tr_x,tr_y,tr_v,br_x,br_y,br_v,bl_x,bl_y,bl_v]
[O, c_x, c_y, w, h, tl_{[x,y,v]}, tr_{[x,y,v]}, br_{[x,y,v]}, bl_{[x,y,v]}] \in \mathbf{R}^{21}
\end{align}
where $O$ is the gate class; $c_x, c_y, w, h \in [0,1]$ are the normalized bounding box center and size; and $tl,tr,br,bl$ represent the top-left, top-right, bottom-right, and bottom-left corners, each with normalized coordinates ($x,y$) and a visibility flag $v$.

The model was trained in a supervised manner on a dataset of 2000 manually labeled images collected from various tracks and environments, as shown in Figure \ref{fig:gate_dataset}. 
Training utilized losses for class, bounding box regression, keypoint regression, and keypoint confidence\cite{yolov8_ultralytics}. 
Data augmentation included randomized color transformations (contrast, brightness), motion blur, Gaussian blur, MixUp, and Mosaic augmentations to improve robustness.
The model was trained on a desktop PC with an NVIDIA RTX 4090 GPU using PyTorch and subsequently converted to TensorRT 8.5.2 modlwith CUDA 11.4 and FP16 precision for deployment on the onboard Jetson Orin NX computer. 
The model has 11.6 million parameters. 
The measured total inference time on the onboard compute is approximately 16.1 ms, corresponding to a frequency of 62 hz. 
The training dataset was incrementally augmented during the competition preparation phase, using COCO pre-trained weights for initialization. 
Images containing no gates were also added to reduce false-positive detections.

% \vspace{-18pt}
\subsection{State Estimation}
\label{sec:state_estimation}

\subsubsection{Gate Association}
To utilize gate detections for pose estimation, detected gates must be associated with the known gate map. 
This was achieved by projecting the 3D gate map onto the 2D image plane using the current estimated drone pose.
Let the set of gate detections be $\mathbf{S}=\{S_{0},...,S_{M-1}\}$, with $S_i=[\mathbf{p}^\mathcal{I}_{i,0}, \mathbf{p}^\mathcal{I}_{i,1}, ..., \mathbf{p}^\mathcal{I}_{i,K-1}]$ containing the $K=4$ rectified pixel coordinates $\mathbf{p}^\mathcal{I}_{ik}=[x^\mathcal{I}_{ik}, y^\mathcal{I}_{ik}]$ of the corners for the $i$-th detection.
Let the gate map be $\mathbf{G}=\{G_{0},...,G_{N-1}\}$, where each $G_{j}=[\mathbf{p}^\mathcal{W}_{j,k}, \mathbf{p}^\mathcal{W}_{j,k+1}, ..., \mathbf{p}^\mathcal{W}_{j,K-1}]$ contains the 3D coordinates $\mathbf{p}^\mathcal{W}_{j,k}=[x^\mathcal{W}_{j,k}, y^\mathcal{W}_{j,k}, z^\mathcal{W}_{j,k}]$ of the corners for the $j$-th map gate in the world frame $\mathcal{W}$.

Given the estimated camera pose ($\mathbf{p}_{\mathcal{C}}^\mathcal{W}$ and $\mathbf{R}^\mathcal{WC}$ being the transformation between the world frame $\mathcal{W}$ and camera frame $\mathcal{C}$), the projection of a 3D map corner $\mathbf{p}^\mathcal{W}_{j,k}$ into the image plane was calculated as:
\begin{equation}
    \begin{split}
        \mathbf{p}^\mathcal{C}_{j,k} = \mathbf{R}^{\mathcal{W}\mathcal{C}} \left( \mathbf{p}^\mathcal{W}_{j,k} - \mathbf{p}_{\mathcal{C}}^{\mathcal{W}} \right), \\
        \mathbf{p}^\mathcal{I}_{j,k} = \frac{1}{\lfloor \mathbf{p}^\mathcal{C}_{j,k} \rfloor _z} \begin{bmatrix} f_x & 0 & c_x \\ 0 & f_y & c_y \end{bmatrix} \mathbf{p}^\mathcal{C}_{j,k},
    \end{split}
\end{equation}
where $\mathbf{p}^\mathcal{C}_{j,k}$ is the corner point in the camera frame, $\lfloor \cdot \rfloor_z$ denotes the z-component, and ($f_x, f_y, c_x, c_y$) are the camera intrinsic parameters.

The association between a map gate $G_{j}$ and a detection $S_{i}$ was determined by minimizing a cost function.
% While reprojection error $e_{repj}$ is commonly used, it can be ambiguous when multiple gates project to similar image locations.
\begin{align}
     e_{repj}(G_j, S_i) = \sum_{k=0}^{K-1} \| p_{j,k}^{\mathcal{I}} - p_{i,k}^{\mathcal{I}} \|^2
\end{align}

Gate correspondences were found by solving an optimal assignment problem utilizing the Hungarian algorithm~\cite{kuhn1955hungarian} to minimize the cost over all potential pairings:
\begin{align}
    \min \sum_{(i,j)} e_{repj}(G_j, S_i).
\end{align}

\subsubsection{Gate-based Drone's Pose Estimation}
For each successfully associated pair $(G_j, S_i)$, we estimated the camera's pose relative to the gate using the Perspective-n-Point (PnP) algorithm. 
We established 3D-2D correspondences between the known 3D corners of the map gate $G_j$ and the detected 2D corners of the detection $S_i$.
We used OpenCV's $\operatorname{SolvePnP}$ function with the $\operatorname{SOLVEPNP\_IPPE\_SQUARE}$ method, which is optimized for planar objects like the square racing gates.
This yielded an estimate of the camera's pose in the world frame based on this single gate detection:
\begin{align}
    (p_{\mathcal{C}, S_i}^W, R_{\mathcal{C}, S_i}^{\mathcal{WC}}) := \operatorname{SolvePnP}(G_j, S_i),
\end{align}
where $(p_{\mathcal{C}, S_i}^\mathcal{W}, R_{\mathcal{C}, S_i}^{\mathcal{WC}})$ is the camera pose given detection measurement $S_i$.

However, given noisy corner detections, the resulting orientation estimate $R_{\mathcal{C}, S_i}^{WC}$ is often noisy and less reliable than the orientation provided by the VIO system. 
Therefore, following~\cite{kaufmann2023champion}, we discarded the PnP rotation estimate and only use the position estimate $p_{\mathcal{C}, S_i}^W$ as a measurement update for the kalman filter to correct the translational drift.

\begin{figure}[t!]
    \centering
    \includegraphics[width=1.0\columnwidth]{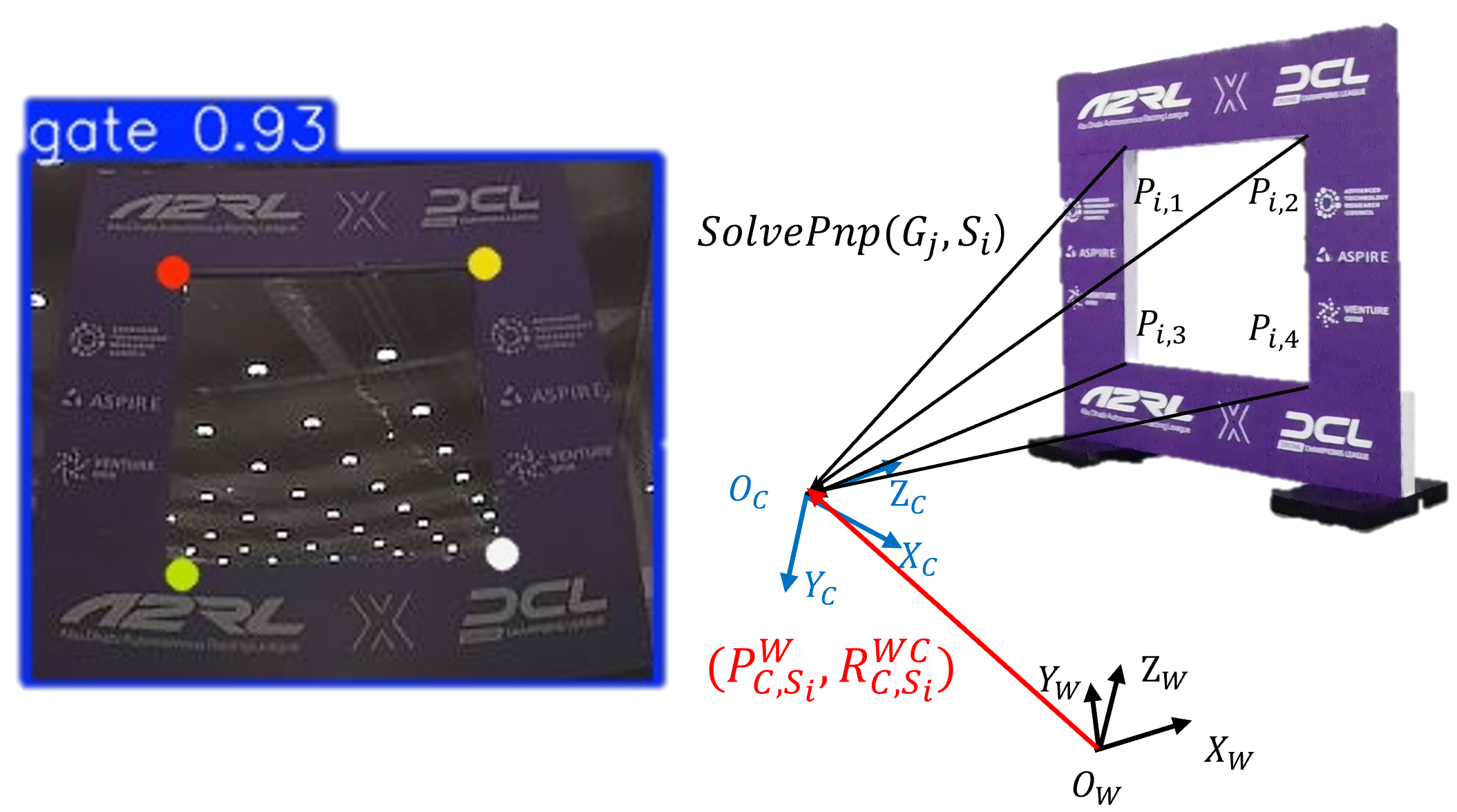}
    \caption{Example of the gate detection result (left) and its gate-based drone pose estimation using perspective-n-point algorithm (right).}
    \label{fig:gate_pose}
\end{figure}

\subsubsection{Detection Filtering}
Gate detections can be unreliable if the gate is too close, too far, heavily tilted relative to the camera, or partially occluded. 
To enhance the robustness of the state estimate, we applied filtering criteria to the associated detections ($G_j, S_i$) before using their PnP position estimate $p_{\mathcal{C}, S_i}^W$ in the kalman filter.

\textbf{Distance Filtering:} Gates very close to the camera can suffer from perspective distortion affecting corner accuracy, while distant gates yield low-resolution detections. 
We computed the distance between the estimated camera position $p_{\mathcal{C}, S_i}^W$ and the known center of the associated map gate $p_{G_j}^W$: $d_{i} =||p_{\mathcal{C}, S_i}^W - p_{G_j}^W||$. 
Detections are discarded if $d_{i} < \tau_{d,min}$ or $d_{i} > \tau_{d,max}$.
We set $\tau_{d,min} = 1$ m and $\tau_{d,max} = 13$ m.

\textbf{Aspect Ratio Filtering:} Significant tilting of the gate relative to the camera skews its appearance, degrading corner detection accuracy. 
We calculated the skew ratio of the detection's bounding box (w,h): $a_i = \operatorname{max}(w/h,\enspace h/w)$. If $a_i > \tau_{ratio}$, the detection is discarded.
where $a_i$ is the skew ratio of $S_i$ and $\tau_{ratio}$ is the threshold.
We set $\tau_{ratio}=2$ to filter out highly skewed gates.

\textbf{Occlusion Filtering:} Gates can be partially occluded by closer gates, leading to incorrect corner detections even if the detector reports high confidence. 
We filtered detections $S_i$ potentially occluded by another detection $S_j$.
A detection $S_i$ is considered occluded and discarded if there exists another detection $S_j$ such that both conditions below are met:
\begin{align}
   \text{visual proximity:} &\quad \operatorname{\Delta p}(S_i,S_j) < \tau_{bbox}, \\
   \text{relative size:} &\quad \frac{A(S_j)}{A(S_i)} > \tau_{area},
\end{align}
where $\operatorname{\Delta p}(S_i,S_j)$ measures the minimum pixel distance between the bounding boxes of $S_i$ and $S_j$. 
A small distance indicates the visual overlap or closeness.
The relative size measures if detection $S_j$ is significantly larger in area than $S_i$ scaled by $\tau_{area}$, which implies $S_j$ corresponds to a close gate that might be occluding $S_i$. 
This combination helps distinguish true occlusion from intentional close configurations like double gates, where the detections in close proximity but the relative size is close to 1.
We set $\tau_{bbox} = 20$px and $\tau_{area} = 1.2$.

\subsubsection{Drift Correction via Kalman Filter}
\label{sec:drift_correction}
The final state estimate integrated the high-frequency but drift-prone VIO output with the less frequent but globally accurate gate-based position measurements using a Kalman Filter (KF)\cite{kaufmann2023champion}. 
The filter aimed primarily to correct the translational drift inherent in VIO during long or aggressive flights.
The filter estimated the drift state rather than the full drone state.
The state vector is $\mathbf{x}_d=[\mathbf{p}_d^\top, \mathbf{v}_d^\top]^\top \in \mathbb{R}^6$, where $\mathbf{p}_d$ is the position drift and $\mathbf{v}_d$ is the velocity drift.

\textbf{Prediction Step:} The drift state was predicted forward using a simple constant velocity model, assuming drift evolves relatively slowly between measurements:
\begin{align}
    \mathbf{x}_{d,k+1}^- = \mathbf{F} \mathbf{x}_{d,k}^+ \enspace, \enspace \enspace \enspace \enspace \mathbf{P}_{k+1}^- = \mathbf{F} \mathbf{P}_k^+ \mathbf{F}^\top + \mathbf{Q}
\end{align}
where $\mathbf{F}$ and $\mathbf{Q}$ are the state transition matrix and process noise covariance, respectively. $F$ and $Q$ are defined below:
\begin{align}
    \mathbf{F} = \begin{bmatrix} \mathbf{I}_{3\times3} & \Delta t \mathbf{I}_{3\times3} \\ 0_{3\times3} & \mathbf{I}_{3\times3} \end{bmatrix}, \enspace
    \mathbf{Q} = \begin{bmatrix} \sigma_p \mathbf{I}_{3\times3} & \mathbf{0}_{3\times3} \\ \mathbf{0}_{3\times3} & \sigma_v \mathbf{I}_{3\times3} \end{bmatrix}
\end{align}
where $\sigma_p$ and $\sigma_v$ are the process noise for the position and velocity components. We set $\sigma_p=0.1$ and $\sigma_v=0.2$.

\textbf{Update Step:} When a valid, filtered gate detection yields a PnP position estimate $\mathbf{z}_k=\mathbf{p}_{S_i}^W$, an update is performed.
The measurement $\mathbf{z}_k$ related the predicted state via the VIO's estimate $\mathbf{p}_{v,k}:\mathbf{z}_k=\mathbf{p}_{v,k}+\mathbf{p}_{d,k}^-$. 
Therefore, the measurement residual is $\mathbf{y}_k=\mathbf{z}_k-\mathbf{p}_{v,k}$. 
With linear measurement model $\mathbf{H}=[\mathbf{I}_{3\times 3} \quad\mathbf{0}_{3\times 3}]$, the kalman filter update is performed by the following:
\begin{equation}
    \begin{split}
    \mathbf{K}_k &= \mathbf{P}_k^- H^\top (\mathbf{H} \mathbf{P}_k^- \mathbf{H}^\top + \mathbf{R})^{-1} \\
    \mathbf{x}_{d,k}^+ &= \mathbf{x}_{d,k}^- + \mathbf{K}_k (y_k - \mathbf{H} \mathbf{x}_{d,k}^-) \\
    \mathbf{P}_k^+ &= (\mathbf{I} - \mathbf{K}_k \mathbf{H}) \mathbf{P}_k^-
    \end{split}
\end{equation}
where $\mathbf{R}=\lambda_r\cdot diag(\sigma_{rx}, \sigma_{ry}, \sigma_{rz}) \in \mathbb{R}^{3\times3}$ is the measurement noise covariance matrix, representing the uncertainty of the PnP position estimate $p_{S_i}^W$.
$\lambda_r>0 \in \mathbb{R}^1$ is a multiplication factor applied to the measurement uncertainty tuned based on the detection quality.
We estimate the measurement uncertainty as the distribution of the pose estimates given 50 pertubed gate detections.

We obtained the final drone state, $\mathbf{X}_{c}$, by correcting the VIO output with the estimated translational drift: $\mathbf{p}_{c} = \mathbf{p}_{v} + \mathbf{p}_d$. 
Other states are taken directly from the VIO output.
\begin{equation}
    \begin{split}
    \mathbf{X}_{c,k} &= \left[\mathbf{p}_{v,k}^\top + \mathbf{p}_{d,k}^\top, \mathbf{q}_{v,k}^\top, \mathbf{v}_{v,k}^\top, \boldsymbol{\omega}_{v,k}^\top,\right]^\top
    \end{split}
\end{equation}
Where $\mathbf{p}_{v,k} \in \mathbb{R}^3$, $\mathbf{q}_{v,k} \in \mathrm{SO}(3)$, $\mathbf{v}_{v,k} \in \mathbb{R}^3$, 
 and $\boldsymbol{\omega}_{v,k} \in \mathbb{R}^3$ are the position, orientation, linear velocity, and angular velocity from the VIO at time $k$.

\subsubsection{Initial Heading Alignment}
\label{sec:init_heading_align}
The VIO system requires motion to initialize, a process we performed manually before the race start. 
We found this process could produce a drifted initial heading, as the VIO would establish its reference frame while the drone was moving unpredictably due to the initialization motion. 
To resolve this, we implemented a one-time heading compensation using gate-based orientation. 
This alignment proved highly effective, as the starting podium consistently offered a clear view of the first gate, ensuring a reliable initial heading for every run.

\subsection{Planning}

\subsubsection{Trajectory Planning}
To generate the race trajectories, our system moved beyond simplistic waypoint-following\cite{Mellinger2011minsnap} by employing the TOGT planner\cite{qin2024togt}. 
Rather than representing gates as single points, TOGT modeled each as a spatial volume, $G_i$, allowing it to find a time-minimal trajectory, $\mathbf{p}(t)$, that passes through the full flyable area. 
The formal optimization problem is expressed as:
\begin{equation}
\begin{aligned}
\min_{\mathbf{p},\,T}\quad & T \\[2pt]
\text{s.\,t.}\quad 
& \mathbf{p}(0)=\mathbf{p}_{\text{start}},\quad \mathbf{p}(T)=\mathbf{p}_{\text{finish}}, \\[4pt]
& \exists\,0<t_1<t_2<\dots<t_L<T \\
  &\text{ such that: } \mathbf{p}(t_i)\in G_i,\; i=1,\dots,L.
\end{aligned}
\end{equation}
The TOGT planner segments the path with waypoints inside each gate, then employs a change-of-variable technique to reformulate the task into an unconstrained optimization problem\cite{wang2022geometrically_traj_optimization}, which is solved efficiently with a gradient-based algorithm\cite{liu1989limited_lbfgs}.
A key practical advantage of TOGT is its computational efficiency over alternatives like CPC\cite{foehn2021CPC}, which enabled a rapid iterative development cycle for trajectory refinement during the limited on-site practice sessions. 
However, the generated solution can be perceptually challenging, as premature or delayed heading turns caused the camera to lose sight of the gate, compromising state estimation. 
To resolve this, we tuned the following parameters: gate's sizes and volumes to limit the flyable area; gate's direction or tilt to generate paths that maintain gate visibility; and body-rates limits.
Furthermore, we also introduce perception-aware planning post trajectory generation to enhance gate-visibility.

\subsubsection{Perception-aware Planning}
Maintaining a clear line of sight to the gates is critical for proposed state estimation module. For autonomous racing, unlike human pilots, the camera could point to any direction as long as a reliable state estimation can be achieved. However, since the arena was surrounded by matte curtains  with no appreciable features, the gates served as the crucial source for reliable localization. Therefore the drone is pointed to the direction of the flight, looking at the gate.
% Traditional time-optimal planners often overlook attitude control, treating it as a byproduct of the trajectory. 
Our approach decouples heading (yaw angle) planning and executes it after trajectory generation.
The core of our general strategy is an anticipatory heading calculation that prevents delayed turns that can lead to a loss of visual lock. 
This was achieved by blending directional cues from the heading toward the current gate, $\psi_{g,i}$, and the heading toward the upcoming gate, $\psi_{g,i+1}$.
This strategy was also inspired by the finding that human pilots fixate their gaze on an upcoming gate as soon as they pass the previous one\cite{pfeiffer2021human}.
To manage this transition smoothly, we employ a distance-based weight, $\lambda_i$, which gradually shifts the drone's focus from the current gate to the next. 
The weight is computed as:
\begin{equation}
\lambda_i = 
\begin{cases}
1 & \text{if } d_{i} < d_{\min} \\
0 & \text{if } d_{i} > d_{\max} \\
\frac{d_{\max} - d_{i}}{d_{\max} - d_{\min}} & \text{otherwise}
\end{cases}
\end{equation}
where $d_{i}$ is the drone's distance to the current gate. 
This weight is used to compute a blended gate-based heading, $\psi_g$.
This heading is then combined with the trajectory's own forward-direction, $\psi_c$, to produce the final desired heading,  
$\psi_{des}$, using weight $\lambda_g$.
\begin{equation}
\begin{split}
    \psi_{g} &= \frac{\lambda_i\psi_{g,i} + \lambda_{i+1}\psi_{g,i+1}}{\lambda_i + \lambda_{i+1}} \\
    \psi_{des} &= \lambda_g\psi_{g} + (1-\lambda_g)\psi_{c}
\end{split}
\end{equation}
where $\psi_{g,i}$ and $\psi_{g,i+i}$ are the heading to gate $G_i$ and $G_{i+1}$, respectively.

While effective for most of the course, this general approach is insufficient for particularly aggressive maneuvers that pose a greater perceptual challenge, such as the split-s maneuver.
The split-s involves a rapid 180$^\circ$ heading reversal, and a standard planner would command a nearly instantaneous heading flip that would saturate the controller.

To handle this special case, we implemented a two-part solution. 
First, to create a safer passage, we added an intermediate virtual gate to the trajectory between the physical split-s gates, pushing the drone further out. 
Second, to execute the turn itself, we proactively manage the heading by incrementally rotating the heading $\psi$ towards the target with a fixed step size, $\Delta\psi_{step}$, ensuring a smooth and controlled inversion that avoids saturating the controller.
\begin{equation}
\psi_{k+1} = \psi_{k} \pm \Delta\psi_{\text{step}}
\end{equation}
where the sign $\pm$ corresponds the turn direction.
 
\subsection{Control} 

\subsubsection{Drone Model}
We first modeled the quadrotor as a rigid body with six degrees of freedom, controlled via collective thrust and body-frame angular velocity. The system state consists of position \( \mathbf{p} \in \mathbb{R}^3 \), velocity \( \mathbf{v} \in \mathbb{R}^3 \), orientation \( \mathbf{q} \in \mathrm{SO}(3) \), and body angular velocity \( \boldsymbol{\omega} \in \mathbb{R}^3 \). The control input is given by the collective thrust \( f \in \mathbb{R}^1 \) and the desired angular velocity \( \boldsymbol{\tau}  \in \mathbb{R}^3 \).

The continuous-time dynamics of the quadrotor are governed by:
\begin{equation}
\begin{split}
    \dot{\mathbf{p}} &= \mathbf{v}    \\
    \dot{\mathbf{v}} &= \frac{1}{m}\mathbf{R}(\mathbf{q})
    \begin{bmatrix} 0 \\ 0 \\ f \end{bmatrix} - \mathbf{g} \\
    \dot{\boldsymbol{\omega}} &= \mathbf{J}^{-1} \left(-\boldsymbol{\omega} \times \mathbf{J}\boldsymbol{\omega} + \boldsymbol{\tau}\right ) \\
    \dot{\mathbf{q}} &= \frac{1}{2}\mathbf{q} \otimes \begin{bmatrix} 0 \\ \boldsymbol{\omega} \end{bmatrix}    
\end{split}
\end{equation}
where $m$ is the vehicle's mass, $\mathbf{g}$ is the gravity vector, and $\mathbf{R}(\mathbf{q})\in \mathbb{R}^{3\times3}$ is the rotation matrix corresponding to the quaternion $\mathbf{q}$. 
The control inputs are the total thrust $f \in \mathbb{R}^1$ and the body-frame torque vector $\boldsymbol{\tau} \in \mathbb{R}^3$. 
The inertia matrix is denoted by $\mathbf{J} = diag(\mathbf{J}_{xx}, \mathbf{J}_{yy}, \mathbf{J}_{zz}) \in \mathbb{R}^{3 \times 3}$, and $\otimes$ represents quaternion multiplication.

\subsubsection{Model Predictive Control}
Our control strategy evolved throughout the championship stages. 
While an initial differential-flatness-based controller\cite{taeyoung2010geometric} was sufficient for qualification stage, we transitioned to a nonlinear Model Predictive Controller (MPC)\cite{foehn2022agilicious} for the more demanding and high-performing final races.
The MPC's predictive nature offered superior tracking of aggressive trajectories. 
This approach allowed for the explicit enforcement of state and input constraints, such as maximum thrust and angular velocity, which was crucial for maintaining stability at the drone's performance limits.
The MPC's cost function was designed to minimize the deviation from the reference trajectory:
\begin{equation}
\begin{aligned}
\min_{u_{0:N-1}} \sum_{k=0}^{N-1} \Big( & \|\mathbf{p}_k - \mathbf{p}_{\text{ref},k}\|^2_{\mathbf{Q}_p} + \|\mathbf{v}_k - \mathbf{v}_{\text{ref},k}\|^2_{\mathbf{Q}_v} \\
& + \|\mathbf{q}_k \ominus \mathbf{q}_{\text{ref},k}\|^2_{\mathbf{Q}_q} + \|\boldsymbol{\omega}_k - \boldsymbol{\omega}_{\text{ref},k}\|^2_{\mathbf{Q}_\omega} \\
& + \|\mathbf{u}_k\|^2_{\mathbf{R}_u} \Big)
\end{aligned}
\label{eq:mpc_problem}
\end{equation}
where $\mathbf{p}_{\text{ref},k}$, $\mathbf{v}_{\text{ref},k}$, $\mathbf{q}_{\text{ref},k}$, and $\boldsymbol{\omega}_{\text{ref},k}$ are the reference trajectory from the planner at time $k$. $\mathbf{u}_k=[f, \boldsymbol{\tau}] \in \mathbb{R}^4$ is the control input. $\mathbf{Q}_p$, $\mathbf{Q}_v$, $\mathbf{Q}_q$, $\mathbf{Q}_{\omega}$ and $\mathbf{R}_u$ are the weights for the tracking errors and control inputs. 

\begin{figure*}[t!] % 'h'ere, 't'op, 'b'ottom, 'p'age of floats
    \centering % Center the figures
    \vspace{-16pt}
     \includegraphics[width=0.97\textwidth]{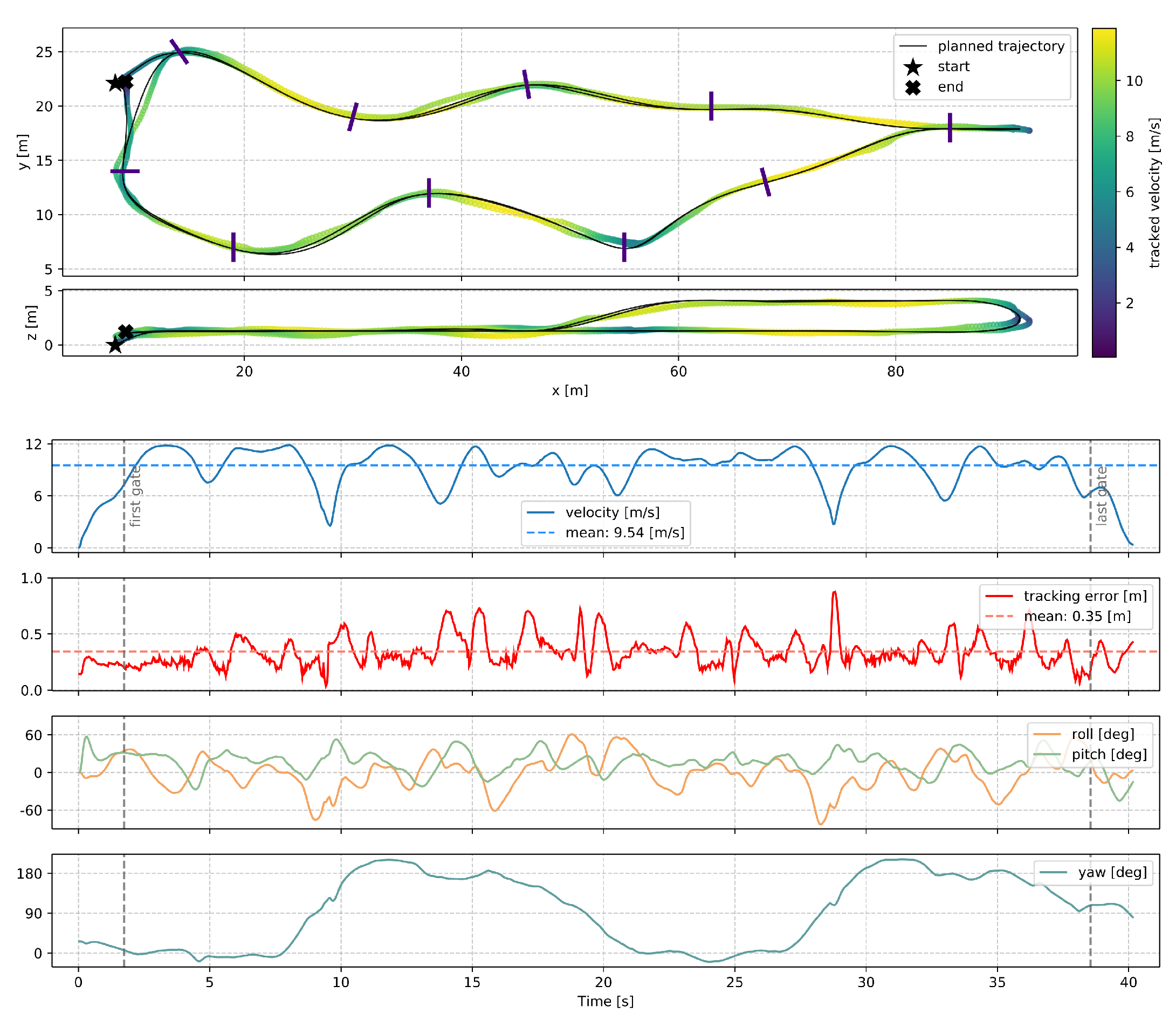}
    \caption{The Third-Place Finishing Run in the AI Grand Challenge Final. The drone closely followed the planned trajectory (top) with an improved mean velocity of 9.54 m/s and a peak speed of approximately 12 m/s. The system maintained impressive precision under race conditions, achieving a mean tracking error of 0.35 m.}
    \label{fig:final_result}
    \vspace{-8pt}
\end{figure*}

The problem in Eq. (\ref{eq:mpc_problem}) was implemented using acados framework\cite{verschueren2022acados} with qpOASES solver, configured with a 1-second prediction horizon and N=20 discretization steps.
The resulting optimal thrust and angular velocity commands were sent to the FC at 200hz, which was then executed using its onboard low-level PID controllers. 
This two-level approach allowed the high-level MPC to focus on trajectory optimization while leveraging the fast and reliable attitude control of the flight controller.

\begin{table}[t!] % Placement specifiers: h-here, t-top, b-bottom, p-page of floats
    \centering % Center the table on the page
    \vspace{-8pt}
    \caption{Summary of Performance Across All Championship Categories.}
    \label{tab:competition_results} % Label for referencing the table
    \setlength{\tabcolsep}{4.8pt}
    \begin{tabular}{ll|c|c|c|c|c} % Column definitions: l-left, c-center
    \hline
    \hline
    \multicolumn{2}{l|}{\multirow{2}{*}{Category}} &\multirow{2}{*}{Rank} &Time &Top Vel &Avg Vel &\multirow{2}{*}{Gates} \\
    {} &{} &{} &[s] &[m/s] &[m/s] &{} \\
    \hline
    \multicolumn{2}{l|}{AI Grand Challenge} & & & &\\
    &Final &3rd &36.8 &12.0 &9.54 &22/22\\
    &Semi-final &3rd &39.0 &12.2 &9.08 &22/22\\
    &Quarter-final &4th &44.3 &10.18 &8.22 &22/22\\
    \hline
    \multicolumn{2}{l|}{AI Drag Race} &2nd &5.4 &16.42 &15.42 &3/3\\
    \hline
    \multicolumn{2}{l|}{AI Multi-Drone Race} &2nd &37.2 &12.2 &9.2 &22/22\\
    \hline
    \multicolumn{2}{l|}{AI vs Human} &top 8 &- &- &- &8/22\\
    \hline
    \hline
    \multicolumn{7}{l}{\parbox{0.95\columnwidth}{
        \begin{itemize}[leftmargin=*]
            \item[-] The time, average velocity and top velocity are measured from the first gate to the last gate.
            \item[-] The values were obtained from our flight logs.
        \end{itemize}
        }}
    \end{tabular}
    \vspace{-8pt}
\end{table}

% \pagebreak
\section{RESULTS}
\label{sec:results}
This section presents a comprehensive analysis of the competition outcomes and evaluation of the proposed system, assessing the performance of its core components.

\begin{figure*}[t!] % 'h'ere, 't'op, 'b'ottom, 'p'age of floats
    \centering % Center the figures
    \vspace{-16pt}
     \includegraphics[width=0.97\textwidth]{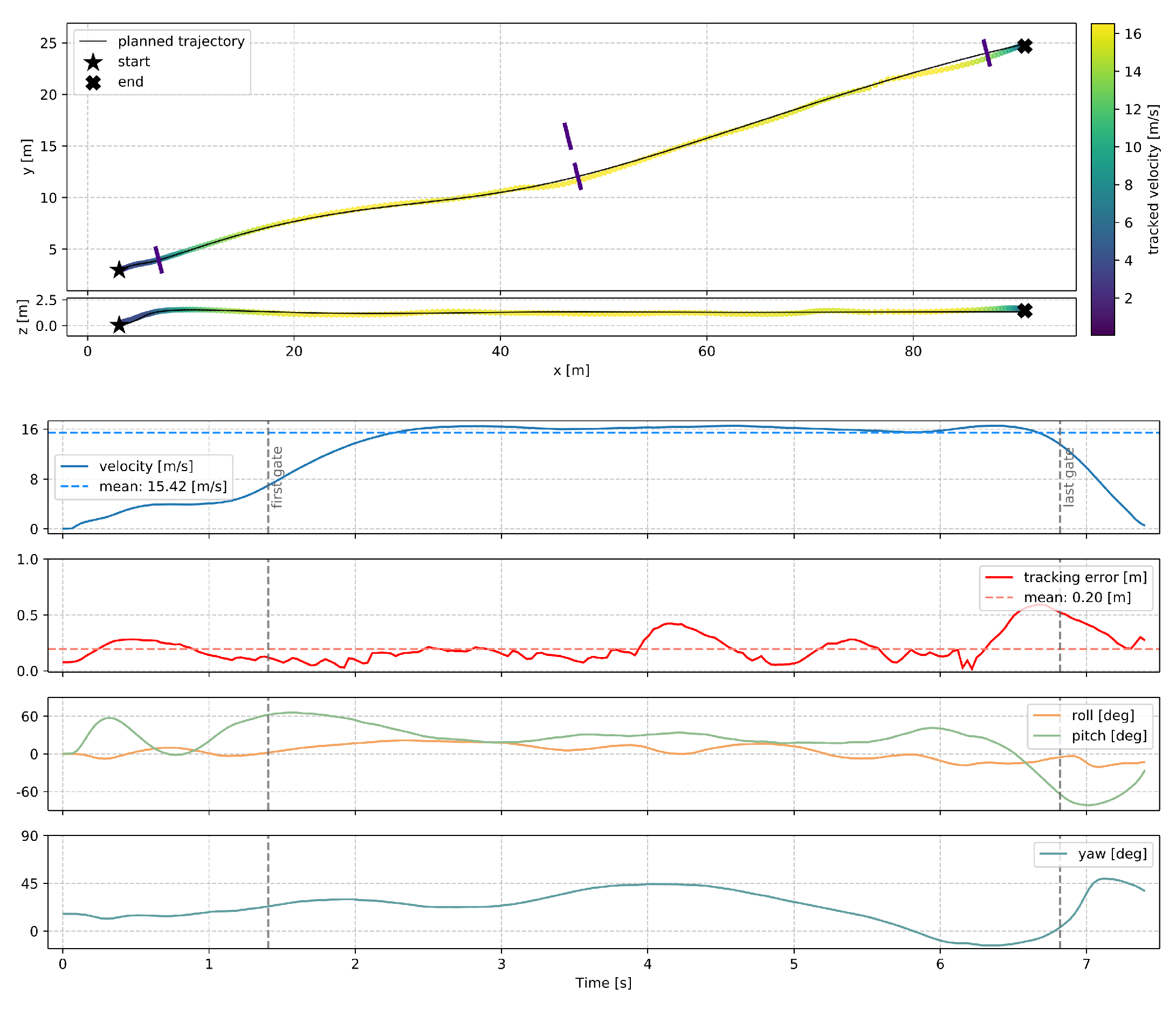}
    \caption{Performance in the AI Drag Race, securing a second-place finish. This figure details the drone's straight-line acceleration and high-speed stability. The system achieved a peak velocity of over 16.4 m/s (approximately 59 km/h) while maintaining exceptional trajectory tracking, as evidenced by a mean tracking error of just 0.20 m.}
    \label{fig:drag_race_result}
    \vspace{-8pt}
\end{figure*}

\subsection{Competition Results and Analysis}
\label{sec:competition_results_analysis}
The A2RL$\times$DCL Championship served as the ultimate validation for our system, testing its performance under intense, real-world racing conditions.
The finals took place at the ADNEC Centre in Abu Dhabi from April 9th to 11th, 2025. 
Over a 10-day period preceding the event, teams were allocated limited practice sessions, typically 30-60 minutes daily, on the official race track. 
The scarcity of practice time---a consequence of the large number of teams, a single available track, and safety protocols preventing simultaneous flights---posed a significant logistical challenge, making the successful completion of a full two-lap run a significant achievement in itself.
We tried to maximally utilize these sessions to collect gate detection data and to fine-tune the parameters of our system modules.

Despite these challenges, our system demonstrated consistent, high-level performance, securing podium finishes in most of the categories. 
As summarized in Table \ref{tab:competition_results}, this included a third-place finish in the AI Grand Challenge, second place in the AI Drag Race, and second place in the AI Multi-Drone Race. 
The following subsections provide a detailed analysis of our performance and the strategies employed in each category.

\subsubsection{AI Grand Challenge}
As the primary category of the championship, the AI Grand Challenge required drones to autonomously complete a two-lap time trial on a 170-meter, 11-gate course with total of 22 gates. 
The competition was structured in three phases: Quarter-final (14 teams), Semi-final (top 8 teams), and Final (top 4 teams). 
In each phase, teams were given two 15-minute slots, with only the best result being recorded for ranking. 
This system incentivized teams to achieve their fastest times early, leading to a high-risk, high-reward environment with numerous crashes.

Our system successfully navigated the course, demonstrating the robustness and effectiveness of our integrated architecture. 
Figure \ref{fig:final_result} provides a detailed analysis of our semifinal run, which secured our qualification for the final. 
The trajectory plots show the drone closely tracking the pre-planned optimal path, maintaining a low mean tracking error of just 0.35 m. 
The system achieved a mean velocity of 9.54 m/s and a peak of 12 m/s. This performance validated our state estimation pipeline, where gate-based corrections effectively mitigated VIO drift, and confirmed the MPC's ability to execute an aggressive trajectory with precision.

In the second slot of the final, we attempted to further improve performance by setting the peak reference velocity to 14 m/s, an untested configuration. 
Initial runs were slower than expected due to a miss-configured controller sampling time. 
Once corrected, the drone achieved the target velocity and completed the first lap one second faster than our podium-finishing run; however, it failed to complete the second lap, as it crashed into the 12th gate during a sharp-turn maneuver. 
During this faster attempt, the drone achieved an average velocity of 10.4 m/s and a peak of 13.5 m/s, but at the cost of a significantly increased mean tracking error of 0.46 m.
This attempt showed that, while our system had achieved competitive performances in this competition, there is still significant room for improvement that can be made to further push the performance.

\subsubsection{AI Drag Race}
The AI Drag Race was a test of straight-line speed and stability, challenging drones to accelerate over a 90-meter straight track. 
While seemingly simpler than the Grand Challenge, this format introduced a unique and critical state-estimation problem. 
With gates positioned far apart, one at the start, two in the middle, and one at the end of the track, the drone had to fly for long stretches without the aid of gate-based drift correction, relying solely on its VIO system for most of the flight. 
This created a high risk of accumulated drift, as our system parameters were primarily tuned for the more gate-dense main track. 
Furthermore, with only one official race attempt without pre-race trial permitted, there was no margin for error in system setup or execution.

Our strategy was to minimize the time spent flying "blind" between gates. 
We planned an aggressive trajectory with a maximum target speed of 16.5 m/s. 
The rationale was that a faster flight would reduce the time available for VIO drift to accumulate, giving the system a better chance of acquiring the next gate for a corrective measurement. 
This approach required absolute confidence in our controller's ability to maintain stability and tracking accuracy at the edge of its performance envelope.

Our system demonstrated exceptional performance, securing a second-place finish. 
As detailed in Figure \ref{fig:drag_race_result}, the drone precisely tracked the planned trajectory, achieving a peak velocity of over 16.42 m/s and an average velocity of 15.42 m/s.
Crucially, the system maintained remarkable stability even at these speeds, evidenced by a mean tracking error of just 0.20 m. 
The attitude plots confirmed the aggressive forward pitch required for high acceleration, while the minimal roll and steady yaw indicate a highly stable flight path.

\subsubsection{AI Multi-Drone Race}
The top eight teams from the AI Grand Challenge advanced to the multi-drone race. Based on the final rankings, teams were split into two groups of four to compete simultaneously. 
As the third-place finisher, our team was placed in the first group, competing against top contenders.
The primary challenge for all teams was the lack of dedicated collision avoidance systems. 
Our strategy was to utilize our proven AI Grand Challenge configuration and rely on the inherent speed differences between drones to maintain safe separation on the track.

At the start of the race, the flight order mirrored the Grand Challenge results.
The race dynamics quickly highlighted the risks of multi-agent navigation without avoidance capabilities as two of the drones crashed and unable to continue the race. 
This incident underscored the complexity of close-proximity racing in the multi-drone category. 
In the end, only one other team and the our system completed the race.
Our system secured a second-place finish with a race time of 37.2 seconds, averaging 9.2 m/s and reaching a peak speed of 12.2 m/s.
Note that on the second group, there was no race finisher, highlighting the difficulty of maintaining a reliable system in a competitive setting.

\subsection{Perception and State-Estimation}

\begin{table}[t!] % Placement specifiers: h-here, t-top, b-bottom, p-page of floats
    \centering % Center the table on the page
    % \vspace{-8pt}
    \caption{Detection performance comparison.}
    \label{tab:detection_comparison} % Label for referencing the table
    \setlength{\tabcolsep}{4.4pt}
    \begin{tabular}{cc|cc|cc} % Column definitions: l-left, c-center
        \hline\hline % Double horizontal line at the top
        \multirow{2}{*}{Size} &\multirow{2}{*}{Model} &Bbox &Kpts &t$_{infer}$ &t$_{total}$\\
         & &mAP &mAP &[ms] &[ms]\\
        \hline % Single horizontal line separating header from data
        \multirow{3}{*}{480px} 
        &YOLOv8n &0.861 &0.959 &4.8$\pm$0.78 &8.7$\pm$0.83 \\
        &YOLOv8s &0.872 &0.966 &6.7$\pm$0.85 &10.6$\pm$0.87 \\
        &YOLOv8m &0.878 &0.965 &12.5$\pm$1.06 &16.4$\pm$1.08 \\
        \hline
        \multirow{3}{*}{640px}
        &YOLOv8n &0.865 &0.959 &7.0$\pm$1.09 &12.6$\pm$1.10\\
        &YOLOv8s &0.877 &0.971 &10.5$\pm$1.33 &16.1$\pm$1.35 \\
        &YOLOv8m &0.885 &0.972 &19.7$\pm$1.16 &25.2$\pm$1.17\\
        
        \hline
        \hline % Double horizontal line at the bottom
        \multicolumn{6}{l}{\parbox{0.8\columnwidth}{
        \begin{itemize}[leftmargin=*]
            \item[-] t$_{infer}$ is the GPU-inference only time, while t$_{total}$ includes the pre- and post-processing time.
        \end{itemize}
        % - 
        }}
    \end{tabular}
    % \vspace{-8pt}
\end{table}

\begin{figure*}[t!]
    \centering
    % \vspace{-12pt}
    \includegraphics[width=\textwidth]{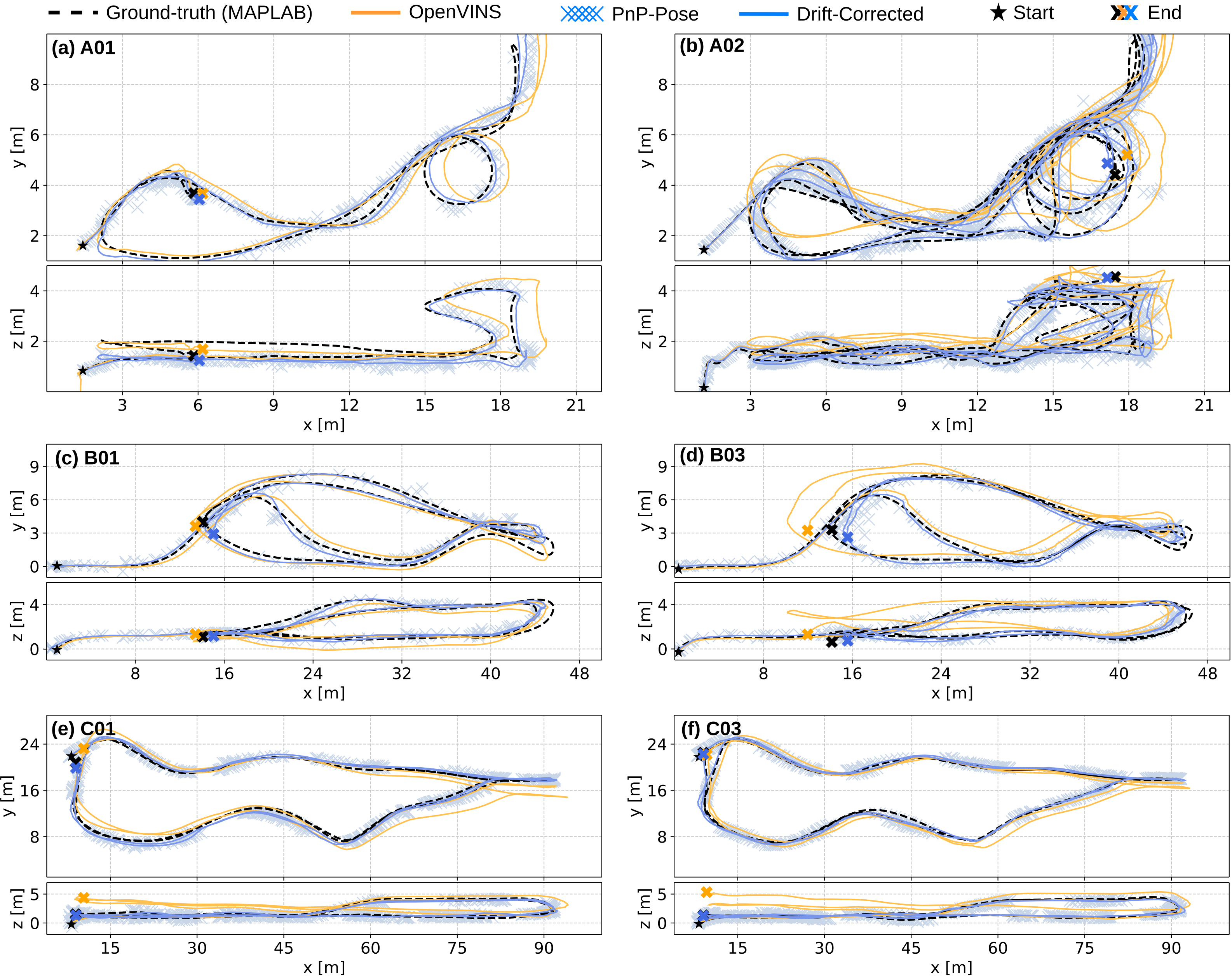}
    \caption{Comparison of the OpenVINS and final state estimation (Drift-Corrected) on ADR-Comp sequence (a) A01, (b) A02, (c) B01, (d) B03, (e) C01, and (f) C03. PnP-Pose is the gate-based pose estimation. Note that the Drift-Corrected estimation converges to the ground-truth improving the accuracy of OpenVINS baseline.}
    \label{fig:vio_state_estimation}
    \vspace{-8pt}
\end{figure*}

\begin{table*}[t!] % Placement specifiers: h-here, t-top, b-bottom, p-page of floats
    \centering % Center the table on the page
    \caption{OpenVINS and its Drift-Corrected State Estimation on the ADR-Comp Dataset.}
    \label{tab:vio_state_estimation} % Label for referencing the table
    \setlength{\tabcolsep}{3.25pt}
    \begin{tabular}{l|cccc|cccccc|cccccc|cc} % Column definitions: l-left, c-center
        \hline\hline % Double horizontal line at the top
        \multirow{2}{*}{\textbf{Method}} &\multicolumn{2}{c}{\textbf{A01}} &\multicolumn{2}{c}{\textbf{A02}} &\multicolumn{2}{c}{\textbf{B01}} &\multicolumn{2}{c}{\textbf{B02}} &\multicolumn{2}{c}{\textbf{B03}} &\multicolumn{2}{c}{\textbf{C01}} &\multicolumn{2}{c}{\textbf{C02}} &\multicolumn{2}{c}{\textbf{C03}} &\multicolumn{2}{|c}{\textbf{Avg}}\\
        &ATE$_t$ &ATE$_r$ &ATE$_t$ &ATE$_r$ &ATE$_t$ &ATE$_r$ &ATE$_t$ &ATE$_r$
        &ATE$_t$ &ATE$_r$ &ATE$_t$ &ATE$_r$ &ATE$_t$ &ATE$_r$ &ATE$_t$ &ATE$_r$ &ATE$_t$ &ATE$_r$\\
        \hline % Single horizontal line separating header from data
        OpenVINS 
            &{0.50} &\tb{3.00} &0.57 &\tb{1.67} &{0.72} &{1.74} &1.28 &3.76 &1.50 &4.11
            &{1.28} &{7.24} &1.21 &3.90 &1.25 &4.90\ &1.04 &3.79\\
        Drift-Corrected 
            &\tb{0.39} &{3.53} &\tb{0.37} &1.85 &\tb{0.68} &\tb{1.52} &\tb{0.66} &\tb{2.41} &\tb{0.77} &\tb{3.08}            
            &\tb{0.59} &\tb{2.60} &\tb{0.52} &\tb{1.43} &\tb{0.49} &\tb{1.31} &\textbf{0.56} &\textbf{2.22}\\
        
        \hline
        \hline % Double horizontal line at the bottom
        \multicolumn{10}{l}{- The best values are \tb{bold}.}
    \end{tabular}
\end{table*}

\subsubsection{Gate Detection}
The performance of the gate detection module was critical for the state estimator's accuracy. We evaluated several YOLOv8-Pose model variants to find the optimal trade-off between detection accuracy and inference speed on the onboard Jetson Orin NX computer. The results of this comparison are summarized in Table \ref{tab:detection_comparison}.

Our chosen configuration, a YOLOv8s model with a 640$\times$640 pixel inference size, achieved a mean average precision (mAP) of 0.877 for bounding box detection and a highly accurate 0.971 mAP for keypoint localization. 
For deployment, the model was converted to a TensorRT engine with FP16 precision, which was crucial for real-time performance. 
The final system recorded a total processing time, including pre- and post-processing, of approximately 16.1 ms per frame. 
While larger models like YOLOv8m offered an increase in accuracy (0.885 bbox mAP and 0.972 keypoint mAP), their processing time of 25.2 ms approached our 33 ms limit, leaving little margin for variability. 
Conversely, smaller models and lower resolutions, while faster, resulted in a decrease in performance.

\begin{figure*}[t!]
    \centering
    \vspace{-8pt}
    \includegraphics[width=1.0\textwidth]{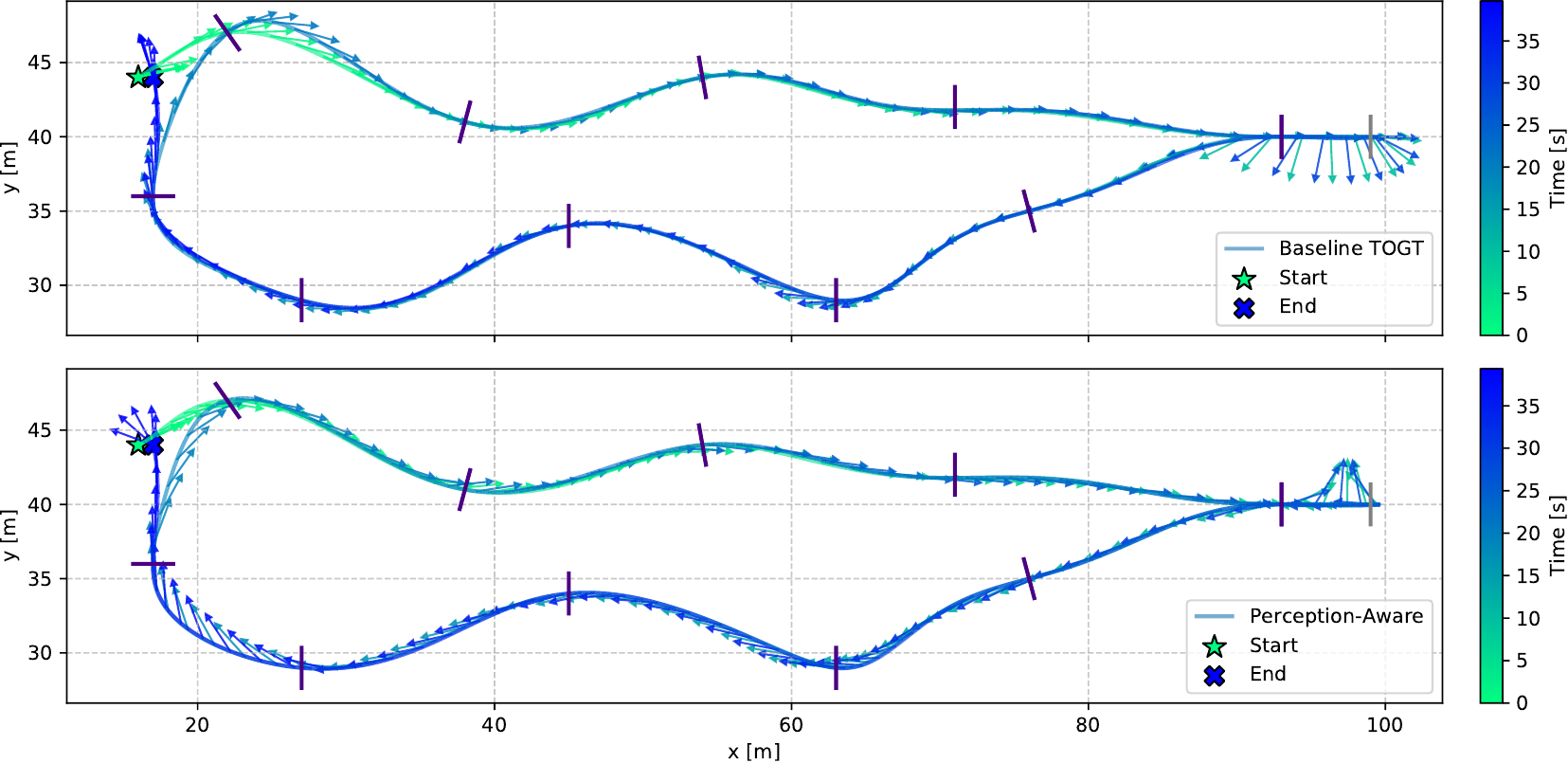}
    \caption{A qualitative comparison of heading planning strategies along the final race trajectory. The arrows indicate the drone's heading at various points. (Top) The baseline TOGT planner\cite{qin2024togt} mostly aligns heading with velocity, causing the camera to lose sight of the next gate during turns. (Bottom) Our perception-aware planner proactively rotates toward the subsequent gate to maximize visibility, ensuring robust state estimation. Gray line is an added virtual gate on split-s maneuver.}
    \label{fig:planning_result_comparison}
    \vspace{-8pt}
\end{figure*}

\subsubsection{VIO and State Estimation}
To evaluate our VIO and state estimation modules, we curated a comprehensive dataset from multiple flights across various track layouts. 
These tracks ranged from the practice session to the final, complex competition course, providing a diverse set of visual conditions.
Specifically, we utilized 2 flights for track A (january practice), 3 for track B (on campuss practice), and 3 for track C (final track).
The ground-truth trajectories are generated using the MAPLAB framework\cite{schneider2018maplab, cramariuc2022maplab}. 
This process involved using an OpenVINS front-end to process the raw visual-inertial data, followed by a full batch optimization of multiple flights simultaneously, to create globally consistent trajectories, even between multiple runs. 
The accuracy of the generated ground truth was qualitatively validated by visually aligning it with the known 3D poses of the race gates. 
Visualization example of the generated ground-truth is depicted in Figure \ref{fig:team_kaist_final_runs}.

The effectiveness of the state estimation pipeline, which fuses OpenVINS with gate detections, was quantified by its ability to mitigate drift over various race tracks. 
The quantitative data in Table \ref{tab:vio_state_estimation} shows a clear and significant improvement in translational accuracy when using the gate-based correction module. 
The drift-corrected states consistently reduced translational error ($ATE_t$) compared to the raw OpenVINS output. 
For instance, on sequence B03, the error was reduced from 1.50 m to 0.77 m, and on sequence C03, it improved from 1.25 m to 0.49 m. 
On average across all evaluated sequences, the fusion strategy reduced the mean translational error from 1.04 m for the OpenVINS baseline to 0.56 m for the final system, providing 45\% relative improvement. 
This confirms the Kalman filter's effectiveness, with the most substantial improvements seen in the longer B and C sequences, which are representative of a full race where drift accumulation is most problematic.

The rotational accuracy (ATE$_r$) showed significant improvement on the B and C sequences, which is attributable to the gate-based initial heading alignment performed prior to the race (Section \ref{sec:impementation}.\ref{sec:state_estimation}.\ref{sec:init_heading_align}). 
This corrective alignment prevents further compounding of rotational error over long flights, which can complicate the correction. 

These quantitative improvements are visually substantiated by the trajectory plots in Figure \ref{fig:vio_state_estimation}.
The trajectory of the OpenVINS visibly drifted away from the ground-truth trajectory over time. 
This deviation was minor on the short track in Figure \ref{fig:vio_state_estimation}a-\ref{fig:vio_state_estimation}b but becomes progressively more severe in the longer B and C sequences. 
In contrast, the final drift-corrected trajectory remains closely aligned with the ground truth across all track layouts, demonstrating the system's ability to successfully correct drift in real time. 
Figure \ref{fig:vio_state_estimation}e-\ref{fig:vio_state_estimation}f, visualizing performance on the longest and most complex track, most clearly shows the necessity of the system; the OpenVINS drifted by several meters, an error that would cause a crash, while the corrected estimate remains highly accurate. 
This combined analysis validates that the state estimation strategy successfully mitigates long-term drift, providing the robust and accurate state awareness required for high-speed autonomous racing.

\subsection{Planning}
To validate our perception-aware planning strategy, we conducted a comparative analysis to quantify its benefits for perceptual stability during aggressive flight. 
We evaluated our complete planner against the baseline TOGT planner\cite{qin2024togt}. 
As we did not push our drone's flight to its physical limits, the evaluation focused on a metric critical for the perception and state-estimation modules: gate visibility.

Gate visibility was measured by projecting the 3D gate corners into 2D pixel coordinates ($u_{i,k}$,$v_{i,k}$). 
We consider a gate is visible if all of its corners lie within the image boundaries of width and height ($W$, $H$) from the image center ($u_0$,$v_0$):
\begin{equation}
\text{vis} = (|u_{i,k}-u_0| \le W/2) \land (|v_{i,k}-v_0| \le H/2)
\end{equation}

\begin{table}[t!]
\vspace{-4pt}
\centering
\caption{Gate visibility comparison with baseline TOGT.}
\label{tab:visibility_scores}
\setlength{\tabcolsep}{6pt}
\begin{tabular}{lc|cc}
\hline \hline
Visibility @ W$\times$H &FOV & TOGT \cite{qin2024togt} & Ours \\
\hline
    vis @ 820$\times$626 px &155$^\circ$$\times$115$^\circ$ &71.36\% &\textbf{80.24\%} \\
    vis @ 614$\times$470 px &120$^\circ$$\times$90$^\circ$ &51.82\% &\textbf{60.19\%} \\
\hline \hline
\multicolumn{4}{l}{\parbox{0.9\columnwidth}{
    \begin{itemize}[leftmargin=*]
        \item[-] FOV is the corresponding evaluation field of view in degree.
        \item[-] The best values are \tb{bold}. 
    \end{itemize}
}}
\end{tabular}
\vspace{-8pt}
\end{table}

The visualization in Figure \ref{fig:planning_result_comparison} highlights the core difference in strategy. 
The baseline TOGT planner, prioritizing time-optimality, mostly keeps the drone's heading aligned with the velocity vector. 
Notably, during sharp turns, there are delayed heading turns (e.g., on gate 1, 5, 8, and 10) which cause the camera to swing away from the trajectory, losing sight of the upcoming gate until the last moment.
In contrast, our perception-aware planner decouples heading from velocity, proactively orienting the drone towards the next gate. 
This anticipatory turn ensures that the target gate enters the field of view earlier and remains visible for a longer duration, providing a more stable stream of measurements for the state estimator.

The quantitative data in Table \ref{tab:visibility_scores} confirms this qualitative observation. 
Our planner achieved a gate visibility of 80.24\% within a wide 155$^\circ$$\times$115$^\circ$ FOV, a substantial improvement over the 71.36\% achieved by the baseline TOGT. 
The benefit is even more pronounced when considering a narrower 120$^\circ$$\times$90$^\circ$ FOV, which represents a more realistic effective field of view. In this case, our method improved the visibility score from 51.82\% to 60.19\%. 
This 16\% relative improvement underscores the effectiveness of our strategy in maintaining perceptual contact, which was crucial for the overall robustness and competitive performance of our system.

\section{DISCUSSION}
\label{sec:discussion}
This section discusses the key takeaways, limitations, and lessons learned from developing and competing with this system.

\subsection{Gate-Centric vs. VIO-Centric State Estimation}
High-speed ADR competitions have produced two distinct state estimation philosophies. 
A \textbf{gate-centric} approach~\cite{li2020adr_vml, de2022mavlab_airr}, relies on gates as the primary source of localization. 
In contrast, a VIO-centric strategy uses VIO as its core, corrected by gate detections\cite{foehn2022alphapilot_uzh, kaufmann2023champion}. 
Both approaches are competitive and have achieved champion-level performance, but each has its own weaknesses.

The gate-centric strategy may struggle in the "perceptual desert" between gates due to high reliance on gate detections features. 
On sparse tracks or during complex maneuvers like the "ladder-up" section, the drone flies blindly and might accumulate significant drift from its motion model. 
A high-risk counter-strategy is to fly faster, reducing the time for drift to accumulate, but this demands exceptional accuracy from the motion model and control algorithm.
This case was exemplified in the January practice session, as most of the teams struggled to pass the "ladder-up" maneuver.
% Our team were able to handle this situation with decent velocity, while another team handled it with significantly higher velocity.
The presented system was able to handle this situation with decent velocity, while another team handled it with significantly higher velocity.
Note that, other teams were not able to perform this maneuver, resulting in the removal of the 'ladder-up' section from the final track.

Conversely, the VIO-centric method's success depends highly on the VIO's quality. 
The Swift platform\cite{kaufmann2023champion} achieved its champions-level performance by using proprietary VIO algorithm on optimized stereo-camera hardware (Intel Realsense T265), while our system is one of the first to achieve reliable results with a monocular camera setup and open-source VIO\cite{geneva2020openvins}. 
By reaching speeds twice as fast as prior VIO-centric systems\cite{foehn2022alphapilot_uzh} in competition settings (16.4 m/s vs. 8 m/s), we demonstrated the viability of this more accessible setup. 
This approach can be further improved with more advanced fusion techniques like MHE\cite{li2020adr_vml} or a tighter integration of gate detections into the VIO optimization/filtering loop\cite{lee2023mins}.

\subsection{Robustifying Gate Detection}
Our approach was to develop a functional detection system under tight time and labor constraints. 
We encountered significant challenges during the final competition, where our detector was initially unreliable. 
This was partly due to a difficult trade-off: we minimized camera exposure time to reduce motion blur for the VIO, but the resulting darker images degraded detection performance. 
Compounding this issue were the major environmental differences between our training tracks and the final venue. 
However, by continually augmenting our dataset and tuning the system on-site, in the end we achieved the reliable performance needed for a podium finish.

This experience underscores the significant room for improvement in gate detection. 
A primary bottleneck is the lack of a large-scale, open-source dataset, which makes it difficult to compare various models for segmentation\cite{de2022mavlab_airr}, keypoint detection\cite{maji_2022_yolo_pose, lu2024rtmo}, or direct 6D pose regression\cite{Pham2021gatenet} objectives for this task. 
Furthermore, manually labeling data from high-speed flights is inherently imprecise due to motion blur and rolling shutter effects, limiting the quality of the labels. 
Future work will explore a larger-scale, photorealistic simulated dataset which could resolve these issues of data scarcity and labeling quality\cite{guerra2019flightgoggles_sim, song2021flightmare_sim, jacinto2024pegasus_sim}, accelerating progress in the field as it has in other areas of computer vision\cite{wang2020tartanair}.

\subsection{Reactive Planning and Adaptive Control}
A key element of our system's success was the tight integration between the planner and the perception system. 
Our approach balanced progress with gate visibility, which was essential for maintaining a stable stream of gate detections for the state estimator. 
This strategy was validated by our results, which show better gate visibility compared to purely time-optimal alternatives\cite{foehn2021CPC, qin2024togt}. 
However, this ad-hoc approach could limit the flight performance as the drone has to fly more conservatively. 
While our decoupled, perception-aware heading controller proved effective, a more principled future direction involves integrating visibility constraints directly into the trajectory optimization framework.
Furthermore, because the entire trajectory is pre-computed, the system is rigid and lacks the ability to adapt in real-time. 
It cannot generate a new path to recover from a significant tracking or state-estimation error, nor can it react strategically to the dynamic actions of opponents in a multi-drone race.

In addition the MPC\cite{foehn2022agilicious} relies on a standard rigid-body dynamics model that does not account for complex aerodynamic effects (e.g., drag, ground effect, downwash) which become significant at high speeds.
Despite having a high-performance tracking, our strategy for the controller and model identification was to treat both as tuning problem, potentially having model mismatch.
Furthermore, different drones may have slightly different dynamic models, and after a crash, the drone's dynamics can change significantly.
While the system was already competitive, more accurate modeling techniques and adaptive control algorithms could potentially resolve this problem~\cite{salzmann2023neural_mpc, gomes2024learningbased_mpc, bauersfeld2021neurobem, wang2023neuralmhe}.
Learning-based control algorithms have also shown promising results\cite{rao2024learnedquad, ferede2024end2end, ferede2025one, eschmann2025raptor}.

\subsection{Future Challenges}
While current systems have been competitive in high-speed navigation on known tracks, the next frontier lies in developing greater intelligence, adaptability, and strategic capability.
There are several future challenges in the field of autonomous drone racing that warrant further investigation. 

\textbf{Multi-Agent Navigation and Strategy.}
A significant unresolved challenge is multi-drone racing that goes beyond simple path-following~\cite{shen2023aggressive_swarm_adr, zhao2025gate_aware_adr}. 
As no team implemented a true collision avoidance or strategic racing algorithm, the crash observed during the multi-drone final underscores this gap. 
The future challenge is not just reactive collision avoidance but proactive, strategic decision-making. 

\textbf{Generalization to Diverse Gate Shapes and Appearances.}
The perception pipelines in most of the ADR systems are highly specialized and specific to square-shaped gates with predefined pattern. 
A significant challenge will be to reliably detect and estimate the pose of diverse gate appearances and shapes (e.g., circles, triangles).

\textbf{Increased Maneuver and Track Complexity.}
Future tracks will undoubtedly demand greater agility by incorporating more difficult configurations. 
Various difficult configurations\cite{getfpv} such as \textit{ladder-up}, \textit{ladder-down}, \textit{corkscrew}, and \textit{reverse split-s} would push the drone to its physical and perceptual limits. 

\textbf{Bridging the Performance Gap to Human Pilots.}
Finally, the "AI vs Human" event clearly demonstrated that a performance gap still exists between top AI teams and elite human pilots.
While the winner is AI drone, the presented system and most of the AI teams' drones were unable to provide a competitive race against human champions, highlighting this disparity. 

\section{CONCLUSION}
\label{sec:Conclusion}

This paper detailed the design, implementation, and competitive performance of the system for the A2RL$\times$DCL Autonomous Drone Championship.
We presented a complete software architecture designed to overcome the core challenge of monocular vision-based racing: the inherent drift of VIO. 
Our solution centered on a robust state estimation pipeline that fused OpenVINS with global position measurements from a YOLO-based gate detector, corrected via a Kalman Filter. 
This was tightly integrated with a perception-aware path planner that balanced time-optimality with the need to maintain gate visibility.
The efficacy of this integrated approach was validated under the high-pressure conditions of the competition. 
Our system demonstrated consistent and reliable performance, securing podium finished in the championship. 
The detailed evaluations and real-world results offer valuable insights and a practical framework for advancing high-speed autonomous navigation with a minimal sensor suite.
\bibliographystyle{ieeetr}
\bibliography{refs}

\vfill\pagebreak

\end{document}